\documentclass[twoside]{article}

\usepackage[accepted]{aistats2025}
\usepackage[round]{natbib}

\usepackage{minitoc}
\usepackage{threeparttable}

\usepackage{wrapfig}
\usepackage[colorlinks=true,
            linkcolor=mydarkblue,
            citecolor=mydarkblue,
            filecolor=mydarkblue,
            urlcolor=mydarkblue]{hyperref}
\usepackage{color,xcolor}         % colors
\hypersetup{
    colorlinks,
    linkcolor={red!50!black},
    citecolor={blue!50!black},
    urlcolor={blue!80!black}
}
\definecolor{mydarkblue}{rgb}{0,0.08,0.45}
\usepackage{mathtools}
\usepackage[page]{appendix}
\usepackage{algorithm}
\usepackage{algpseudocode}

\usepackage{amsmath,amsfonts,amssymb}
\usepackage{subfig}
\usepackage{mathtools}
\usepackage{enumitem}
\usepackage{tikz}
\usepackage{wrapfig}
\usepackage{lipsum,booktabs}
\usepackage{graphicx,scalerel}
\usepackage{subcaption}
\usepackage[amsthm,thmmarks,amsmath]{ntheorem}
\usepackage{thmtools,thm-restate}

\newcommand{\nindep}{\not\perp\!\!\!\perp}

\begin{document}

\doparttoc % Tell to minitoc to generate a toc for the parts
\faketableofcontents % Run a fake tableofcontents command for the partocs



\twocolumn[

\aistatstitle{Nonparametric Factor Analysis and Beyond}

\aistatsauthor{ Yujia Zheng$^{1}$ \And Yang Liu$^{2}$ \And  Jiaxiong Yao$^{2}$ \And Yingyao Hu$^{\dagger,3}$ \And Kun Zhang$^{\dagger,1,4}$ }

\aistatsaddress{ $^1$Carnegie Mellon University \\$^2$International Monetary Fund \\ $^3$Johns Hopkins University \\ $^4$ Mohamed bin Zayed University of Artificial Intelligence }
% \aistatsaddress{ $^1$CMU $^2$IMF  $^3$JHU $^4$MBZUAI }
]

\begin{abstract}
  Nearly all identifiability results in unsupervised representation learning inspired by, e.g., independent component analysis, factor analysis, and causal representation learning, rely on assumptions of additive independent noise or noiseless regimes. In contrast, we study the more general case where noise can take arbitrary forms, depend on latent variables, and be non-invertibly entangled within a nonlinear function. We propose a general framework for identifying latent variables in the nonparametric noisy settings. We first show that, under suitable conditions, the generative model is identifiable up to certain submanifold indeterminacies even in the presence of non-negligible noise. Furthermore, under the structural or distributional variability conditions, we prove that latent variables of the general nonlinear models are identifiable up to trivial indeterminacies. Based on the proposed theoretical framework, we have also developed corresponding estimation methods and validated them in various synthetic and real-world settings. Interestingly, our estimate of the true GDP growth from alternative measurements suggests more insightful information on the economies than official reports. We expect our framework to provide new insight into how both researchers and practitioners deal with latent variables in real-world scenarios.

  % Latent variable models play a pivotal role in scientific research, particularly when key variables are unobserved and must be inferred. While numerous studies have explored the theoretical guarantees for identifying latent variables, most are limited by restrictive parametric assumptions imposed on the latent model, or the noise, or both. 
  
  % In this paper, we propose a general framework for identifying latent variables in nonparametric models, even in the presence of non-negligible noise. We first show that under suitable conditions, the model is identifiable up to certain submanifold indeterminacies. Furthermore, under the structural or distributional variability conditions, we prove that latent variables of general nonlinear models are identifiable up to trivial indeterminacies even in the presence of non-negligible noise.  Based on the proposed theoretical framework, we have also developed corresponding estimation methods and validated them in various synthetic and real-world settings. Interestingly, our estimate of the true GDP growth from alternative measurements suggests more insightful information on the economies than official reports. We expect our framework to provide new insight into how researchers deal with latent variables in real-world scenarios.
\end{abstract}

\section{Introduction}
\label{sec:introduction}
\vspace{-0.3em}

Revealing the hidden process that generates observed data is fundamental to scientific discovery. A typical example is the so-called hidden Markov model, where a series of latent variables are observed with errors in multiple periods under conditional independence. Although machine learning can model intricate patterns in data, it frequently falls short in ensuring that its representations match the true underlying factors driving the observations \citep{locatello2019challenging}. Reliable identification of these latent factors is crucial for unbiased analysis across various fields, such as economics \citep{hu2008identification,hu2017econometrics,schennach2020mismeasured,Hu2025} and psychology \citep{bollen2002latent, marsh2007applications}.

A typical set of approaches to identifying the hidden process underlying data generation have predominantly addressed linear relations between hidden and observed variables, providing strong theoretical backing; see, e.g., \citep{aigner1984latent, comon1994independent, bishop1998latent}. Recent developments, for instance, nonlinear independent component analysis (ICA), have broadened this focus to capture more intricate, nonlinear relationships \citep{hyvarinen1999nonlinear, hyvarinen2024identifiability}. These methods often introduce additional requirements, such as leveraging auxiliary variables \citep{hyvarinen2016unsupervised}, utilizing time-series information \citep{hyvarinen2017nonlinear}, imposing structural assumptions \citep{zhengidentifiability}, or specifying certain functional forms \citep{taleb1999source}. Despite these advances, many models assume a noise-free environment, limiting their effectiveness in practical situations where data is inherently subject to random fluctuations.

Some works have focused on the latent variable models in noisy settings. For instance, classical factor analysis models \citep{reiersol1950identifiability,lawley1962factor, bekker1997generic} can incorporate noise but remain subject to certain limitations. First, like several approaches in noisy ICA and other models \citep{ikeda2000independent, beckmann2004probabilistic, bonhomme2009consistent, khemakhem2020variational}, factor analysis is constrained to handle noise that is specifically \textit{additive} and \textit{independent} of the latent variables. This restricts its flexibility in real-world scenarios where noise may interact with latent representation in more complex ways. Second, factor analysis relies on a fundamentally linear relationship between latent variables and observations or the model that can be reduced to a linear one, limiting its capacity to capture general nonlinearity in the generative processes. Furthermore, even when these assumptions hold, the model's identifiability is usually only guaranteed up to a linear subspace, leaving latent variables partially entangled and preventing a complete recovery of the true underlying factors. These limitations underscore the need for more general frameworks that can handle broader classes of noise and nonlinear relationships while providing stronger identifiability guarantees.

To address those concerns, we establish a general framework for identifying latent variables in nonparametric models with \textit{nonlinear} generating processes based on the so-called Hu-Schennach Theorem, even when confronted with \textit{non-negligible} noise. The generality of both the latent model and the noise allows us to tackle complex nonlinear transformations underlying the data, even when the generative process is \textit{noninvertible} due to the general noise. We first show to what extent the nonparametric factor analysis model is identifiable. Moreover, unlike previous work in factor analysis, our focus extends beyond submanifold identification, demonstrating that all latent variables can be identified, thereby fully disentangling the underlying mixture of generative factors. Specifically, we show that, under standard conditions, such as structural or distributional variability, latent variables of nonparametric models are identifiable up to a permutation and component-wise invertible transformation. We also propose estimation methods to support this identification and validate our results on both synthetic and real-world datasets. Notably, we demonstrate that GDP growth estimates derived from alternative measurements, like Google search trends and nightlight intensity, offer deeper insights into economic conditions than traditional official reports.

\vspace{-0.3em}
\section{Preliminary}
\label{sec:prelim}
\vspace{-0.3em}

We consider a general data-generating process as follows:
\begin{equation} \label{eq:generating}
    X = f(Z, \epsilon),
\end{equation}
where $X = (X_1, X_2,\ldots,X_m) \in \mathcal{X} \subseteq \mathbb{R}^{m}$ denotes observed variables,  $Z = (Z_1, Z_2,\ldots,Z_n) \in \mathcal{Z} \subseteq \mathbb{R}^{n}$ denotes latent variables, and $\epsilon$ denotes noise. Notably, we do \textit{not} require independent noise and thus it is fully possible that $Z \nindep \epsilon$, which is different from most previous work in the literature. Moreover, the function $f$ is generally not invertible, further extending the considered setting.

\textbf{Technical Notations.} \ We use capital letters to stand for a random variable and lower case letters to stand for the realization of a random variable. For example, $p_{V}(v)$ denotes the probability density function of random variable $V$ with realization argument $v$, and $p_{V|U}(v|u)$ denotes the conditional density of $V$ on $U$. The capital letter $P$ denotes the distribution.

Throughout this work, for any matrix $S$, we use $S_{i,:}$ to indicate its $i$-th row and $S_{:,j}$ to indicate its $j$-th column. For any index set $\mathcal{I} \subset \{1, \ldots, a\} \times \{1, \ldots, b\}$, we define $\mathcal{I}_{i,:} \coloneqq \{j \mid (i, j) \in \mathcal{I}\}$ and $\mathcal{I}_{:,j} \coloneqq \{i \mid (i, j) \in \mathcal{I}\}$. Based on this, we define the support of a matrix $S \in \mathbb{R}^{a \times b}$ as $\operatorname{supp}(S)\coloneqq\left\{(i,j) \mid S_{i,j} \neq 0 \right\}$. Similarly, the support of a matrix-valued function $\mathbf{S}(\Theta): \Theta \rightarrow \mathbb{R}^{a \times b}$ is defined as $\operatorname{supp}(\mathbf{S}(\Theta))\coloneqq\left\{(i,j) \mid \exists \theta \in \Theta, \mathbf{S}(\theta)_{i,j} \neq 0 \right\}$. Furthermore, for any subset $\mathcal{S} \subseteq \{1, \ldots, n\}$, we define its subspace $\mathbb{R}_{\mathcal{S}}^n$ as $\mathbb{R}_{\mathcal{S}}^n \coloneqq \{ s \in \mathbb{R}^n \mid s_i = 0, \forall i \notin \mathcal{S} \}$, where $s_i$ is the $i$-th element of the vector $s$. All estimated quantities are denoted using the hat symbol, e.g., $\hat{Z}$ and $\hat{f}$. For ease of reference, we have included a summary of the notation in Appendix \ref{appx:notation}.

\vspace{-0.3em}
\section{Identifiability Theory}
\label{sec:identification}
\vspace{-0.3em}

Suppose that the ideal data for estimating a model consists of a sample of $(X, Z)$, but the researcher only observes $X$. Our objective is to identify the latent variable(s) $Z$ under the most general conditions. We first show how to identify the latent manifold (Section \ref{sec:dist}), and then the identifiability of individual latent variables (Section \ref{sec:multi}). 

\vspace{-0.3em}
\subsection{Distribution Identifiability}
\label{sec:dist}
\vspace{-0.3em}

We assume that a researcher observes the distribution of $\{X_1, X_2,\ldots,X_m\}$ from a random sample. Putting the estimation of the population distribution $P_{X_1, X_2,\ldots,X_m}$ from the random sample aside, we face a key identification challenge: How to determine the distribution $P_{X_1, X_2,\ldots,X_m,Z}$ from the observed distribution $P_{X_1, X_2,\ldots,X_m}$. We first introduce a nonparametric identification result for the hidden distribution.

\begin{restatable}{assumption}{assump0}
\label{assumption 3.0} 
The observed variables $X$ can be split into three parts $\{X_A, X_B, X_C\}$, where variables in each part are conditionally independent of variables in other parts given $Z$.
% There exists a random variable $Z$ with support $\mathcal{Z}$ such that  
% \begin{eqnarray}  \nonumber
% &&p_{X_1, X_2,\ldots,X_m,Z} \\ \nonumber
% &=& p_{X_1|Z}  p_{X_2|Z} \cdot \ldots \cdot p_{X_m|Z} p_{Z}.
% \end{eqnarray}\normalsize
\normalsize 
\end{restatable}

We may consider the observables $(X_1, X_2,\ldots,X_m)$ as measurements of $Z$. Assumption 1 implies the conditional independence structure, which is commonly observed in many real-world scenarios. For instance, symptoms such as fever, cough, and muscle aches may exhibit dependencies but are conditionally independent given the latent cause, such as influenza. Here we leverage \citep{hu2008instrumental} to show the uniqueness of $p(X_1, X_2,\ldots,X_m,Z)$. We make the following assumption.
\begin{restatable}{assumption}{assump1}
\label{assumption 3.1} The joint distribution of $(X_1, X_2,\ldots,X_m, Z)$ admits a bounded density with respect to the product measure of some dominating measure defined on their supports. All marginal and conditional densities are also bounded.
\end{restatable}

Assumption \ref{assumption 3.1} requires the densities to be bounded because the decomposition of linear operators is well-established for bounded linear operators. For unbounded linear operators, the uniqueness of the decomposition is quite challenging, which previous works did not explore. Nevertheless, the support of the densities can be the whole real line, i.e., unbounded. Note that it is possible to transform an unbounded density over a bounded support to a bounded density over an unbounded support, so it can be extended to some cases where densities are unbounded. 

Before introducing more assumptions, we define an integral operator corresponding to $p_{X_A\vert Z}$, which maps $p_{Z}$ over support $\mathcal{Z}$ to $p_{X_A}$ over support $\mathcal{X}_A$. Suppose that we know both $p_{Z}$ and $p_{X_A}$ are bounded and integrable. We define $\mathcal{L}_{b n d}^{1} \left (\mathcal{Z} \right )$ as the set of bounded and integrable functions defined on $\mathcal{Z}$, i.e.,
\begin{eqnarray}\nonumber
&&\mathcal{L}_{b n d}^{1} \left (\mathcal{Z}\right ) \\ \nonumber
&=&\left \{g :\int _{\mathcal{Z}}\left \vert g (z)\right \vert   dz <\infty \;, \;\sup_{z \in \mathcal{Z}}\left \vert g (z)\right \vert  <\infty \right \}. \label{equ 060}
\end{eqnarray}\normalsize
\normalsize
The linear operator can be defined as
\begin{eqnarray}L_{X_A\vert Z} &  : & \mathcal{L}_{b n d}^{1} \left (\mathcal{Z}\right ) \rightarrow \mathcal{L}_{b n d}^{1} \left (\mathcal{X}_A\right ) \label{equ 070} \\
\left (L_{X_A\vert Z} h\right ) \left (x\right ) &  = & \int _{\mathcal{Z}}p_{X_A\vert Z} (x\vert Z)h (Z)  dZ . \nonumber \end{eqnarray}\normalsize  
\normalsize
In order to identify the unknown distributions, we need the observables to be informative so that the following assumptions hold.
\begin{restatable}{assumption}{assump2}
\label{assumption 3.2}The operators $L_{X_A\vert Z}$ and $L_{X_B\vert X_A}$ are injective.\protect\footnote{ $L_{X_B\vert X_A}$ is defined in the same way as $L_{X_A\vert Z}$ in Eq. \eqref{equ 070}.}
\end{restatable}

Assumption \ref{assumption 3.2} intuitively introduces sufficient variation in the densities, which is a mild and common condition in the nonparametric identification literature. It is also equivalent to the completeness of the density over a certain functional space \citep{mattner1993some}.

\begin{restatable}{assumption}{assump3}
\label{assumption 3.3}For all $\overline{z} \neq \widetilde{z}$ in $\mathcal{Z}$, the set $\left \{x_C :p_{X_C\vert Z} \left (x_C\vert \overline{z}\right ) \neq p_{X_C\vert Z} \left (x_C\vert \widetilde{z}\right )\right \}$ has positive probability.
\end{restatable}

Assumption \ref{assumption 3.3} is a generally mild condition to ensure each possible value of the latent variable affects the distribution of observed variables. It is only violated when the conditional distribution is identical for two distinct values of the conditioning variable.

\begin{restatable}{assumption}{assump4}
\label{assumption 3.4}There exists a known functional $M$ such that $M\left [p_{X_A\vert Z} \left ( \cdot \vert Z\right )\right ] =Z$ for all $Z \in \mathcal{Z}$.
\end{restatable}

 The functional $M$ may be the mean, mode, medium, an arbitrary quantile of the probability measure $p_{X_A\vert Z} \left ( \cdot \vert Z\right ) $, or any other properties. The identification result may be summarized as follows:

\begin{restatable}{theorem}{HuSc2008} \label{HuSc2008}
\citep{hu2008instrumental} Under assumptions \ref{assumption 3.0}, \ref{assumption 3.1}, \ref{assumption 3.2}, \ref{assumption 3.3}, and \ref{assumption 3.4}, the joint distribution $p_{X_1, X_2,\ldots,X_m}$ uniquely determines the joint distribution $p_{X_1, X_2,\ldots,X_m,Z}$, which satisfies 
\begin{equation}\label{equ 200}
p_{X_1, X_2,\ldots,X_m,Z} = p_{X_A|Z} p_{X_B|Z} p_{X_C|Z} p_{Z}. 
\end{equation}
\end{restatable}

This identification result implies that if we have three sets of qualified measurements $X_A$, $X_B$ and $X_C$, we are able to provide a consistent estimator of $p_{X_1, X_2,\ldots,X_m,Z}$, or $p_{Z\,|\,X_1, X_2,\ldots,X_m}$, from a sample of $(X_1, X_2,\ldots,X_m)$.  

Notably, Assumption \ref{assumption 3.4} plays a role in determining the order of values. Without this assumption, we can only identify the corresponding submanifold rather than the full distribution. However, as we will demonstrate later, for the purpose of identifying latent variables, recovering the submanifold is sufficient. Therefore, Assumption \ref{assumption 3.4} is not essential for the broader framework.

% This identification result only needs three sets of variables that are conditionally independent across sets. Therefore, the conditional independence may be relaxed to 
% \begin{equation*} 
% p_{X_1, X_2,\ldots,X_m,Z} = p_{X_1|Z} p_{X_2|Z} p_{X_3,\ldots,X_m,Z}. 
% \end{equation*}

% In the remaining discussion, we still use the conditional independence in equation (\ref{equ 200}) because we are interested in the common element $Z$ across all the observables. 

% This identification result implies that if we have qualified measurements $X_1$, $X_2$ and $X_3$, we are able to provide a consistent estimator of $p_{X_1, X_2,\ldots,X_m,Z}$, or $p_{Z\,|\,X_1, X_2,\ldots,X_m}$, from a sample of $(X_1, X_2,\ldots,X_m)$. 

\vspace{-0.3em}
\subsection{Identifiability of Latent Variables}
\label{sec:multi}
\vspace{-0.3em}

In the previous section, we know that the identifiability of distribution can be guaranteed in a nonparametric manner. However, identifying the distribution is often not sufficient for many practical applications. In many scenarios, we need to recover the individual latent components to gain deeper insights into the underlying processes. For instance, in understanding complex systems like economic indicators or biological mechanisms, it is crucial not only to know the latent distribution but also to disentangle the specific factors leading to the observations. This component-wise identification allows for a more granular understanding, enabling targeted interventions, improved interpretations, and more precise inference. Therefore, it becomes necessary to go beyond distributional identifiability and focus on recovering the individual latent components.

We first propose the following theorem for identifying the submanifold of the latent variables.

\begin{restatable}{theorem}{nfsuff}\label{thm:nf_suff}
    Consider two models $\theta = (f,p_Z,p_{\epsilon})$ and $\hat{\theta} = (\hat{f},p_{\hat{Z}},p_{\hat{\epsilon}})$ following the process in Section \ref{sec:prelim}, under Assumptions \ref{assumption 3.0}, \ref{assumption 3.1}, \ref{assumption 3.2}, and \ref{assumption 3.3}, there exists an invertible function $h$ such that 
    \begin{equation*}
        p(x;\theta) = p(x;\hat{\theta}) \implies \hat{Z} = h(Z).
    \end{equation*}
\end{restatable}

The proof is in Appendix \ref{sec:proof_nf_suff}. Different from Theorem \ref{HuSc2008}, we remove Assumption \ref{assumption 3.4} to minimize the reliance on prior knowledge. As a trade-off, for now, we can only identify $\mathbf{Z}$ up to a submanifold instead of pinning down the whole distribution. But we will soon show that, under appropriate conditions, the submanifold identification leads to the desired component-wise identification.

We also have the following results to ensure that the dimension of the recovered unobserved variable cannot be further reduced under continuity, with their proofs in Appendix \ref{sec:proof:lem_reduc1d} and \ref{sec:proof:lem_reduckd}.

\begin{restatable}{lemma}{ReduceoneD}\label{lem:reduc1d}
    A one-to-one function $g: \mathbb{R}^n \rightarrow \mathbb{R}$ cannot be continuous. Therefore, the dimensionality of $\mathbb{R}^n$ cannot be reduced to that of $\mathbb{R}$ under the continuity. 
\end{restatable}

\begin{restatable}{lemma}{ReduceKD}\label{lem:reduckd}
    A one-to-one function $g: \mathbb{R}^n \rightarrow \mathbb{R}^k$ with $k<n$ cannot be continuous. Therefore, the dimensionality of $\mathbb{R}^n$ cannot be reduced to that of $\mathbb{R}^k$ under the continuity.  
\end{restatable}

These results have important implications for our identification strategy. They imply that the unobserved variable $Z$ must retain its intrinsic dimensionality when being recovered from observed data under continuity assumptions. In other words, we cannot hope to represent a higher-dimensional unobserved variable using a lower-dimensional observed counterpart without losing continuity or injectivity. This reinforces the necessity of our approach in maintaining the dimensionality of $Z$ during estimation to ensure accurate and unique recovery of its realizations, which is especially essential for cases with multivariate latent variables.

Based on these, we are now ready to move beyond the submanifold identifiability to the component identifiability of the latent variables, under appropriate conditions on the connective structure between $Z$ and $X$, or the changeability of the latent distribution. For simplicity, we denote the support of the Jacobian $J_f$ as $\mathcal{F}$, i.e., $\mathcal{F} = \operatorname{supp}(J_f)$, where $J_f$ is the derivative of $f$ with respect to $Z$. Additionally, let $\mathcal{T}$ represent the set of matrices with the same support as $\mathbf{T}$ in the equation $J_{\hat{f}} = \mathbf{T} J_f$, where $\mathbf{T}$ is a matrix-valued function.

\begin{restatable}{theorem}{sparsity}\label{thm:sparsity}
    Consider two models $\theta = (f,p_Z,p_{\epsilon})$ and $\hat{\theta} = (\hat{f},p_{\hat{Z}},p_{\hat{\epsilon}})$ following the process in Section \ref{sec:prelim}. In addition to the assumptions in Theorem \ref{thm:nf_suff}, suppose $|\hat{\mathcal{F}}| \leq |\mathcal{F}|$ and the following assumptions hold:
    \begin{enumerate}[label=\roman*.,ref=\roman*]
      \item The density $p_Z$ is positive and smooth.

      \item For each $i \in \{1, \ldots, n\}$, there exists a set of points $ \{z^{(\ell)}\}_{\ell=1}^{|\mathcal{F}_{i,:}|}$ and a matrix $\mathrm{T} \in  \mathcal{T}$ s.t. $\operatorname{span}\{J_f(z^{(\ell)})_{i,:}\}_{\ell=1}^{|\mathcal{F}_{i,:}|} = \mathbb{R}_{\mathcal{F}_{i,:}}^{n}$ and $\left[ {J_f(z^{(\ell)})}\mathrm{T} \right]_{i,:} \in \mathbb{R}_{\hat{\mathcal{F}}_{i,:}}^{n}.$
      \label{assum:nondeg}
    
      \item (\underline{Structural Variability}) For each $k \in \{1, \ldots, n\}$, there exists $\mathcal{C}_{k}$ s.t. $\bigcap_{i \in \mathcal{C}_{k}} \mathcal{F}_{i, :}=\{k\}.$
      \label{assum:sparsity}
    \end{enumerate}
    Then there exists a component-wise invertible function $h$ and a permutation $\pi$ such that, $\forall i \in \{1, \ldots, n\}$,
    \begin{align*}
        p(x;\theta) = p(x;\hat{\theta}) \implies \hat{z}_i = h_i(\pi(z_i)).
    \end{align*}
\end{restatable}

The proof is in Appendix \ref{sec:proof:thm_sparsity}. The structural variability (Assumption \ref{assum:sparsity}) has been introduced in \citep{zhengidentifiability}. However, the identifiability results in \citep{zhengidentifiability} are limited to deterministic transformations, thus requiring the generative process to be a diffeomorphism without any noise. In Theorem \ref{thm:sparsity}, we prove that, even in the presence of non-negligible noise, we can still identify the latent variables up to the same component-wise indeterminacy as that in the previous results.

Intuitively, Assumption \ref{assum:nondeg} avoids some unlikely cases where the samples are all from a very small population that spans only a degenerate submanifold. Therefore, it is always almost satisfied in the considered asymptotic case. Assumption \ref{assum:sparsity} requires sufficient structural diversity on the connective structure between latent and observed variables. For example, in a biomedical context, consider latent variables representing different underlying health conditions or genetic factors, with observed variables such as blood pressure, cholesterol levels, and glucose levels. It is unlikely that each health condition would impact exactly the same set of clinical measurements. A latent condition related to cardiovascular health might influence blood pressure and heart rate, while another condition related to metabolic health might affect glucose and cholesterol levels. Even if some overlap exists, the pattern of which latent variables affect which observed variables differs, ensuring distinct dependency structures. 

Moreover, since the structural variability assumption only requires a subset of observed variables to meet the condition, it is generally satisfied when the number of observed variables exceeds the number of latent variables, as shown in \citep{zheng2023generalizing}. When the number of observed variables is greater than that of the latent variables, even if the current structure does not initially meet the condition, additional measurements can be introduced to achieve the required variability. Therefore, this provides a practical way to manually ensure that the assumption is met, which is particularly valuable given that real-world generative processes are often unknown, and most identifiability conditions in the literature are not directly testable.

It might be worth noting that, the structural sparsity assumption implicitly requires that the latent variables are independent of one another. This aligns with recent research in nonlinear ICA, which assumes independence among latent variables along with additional conditions to achieve identifiability. 
Following the spirit of the seminal foundations laid by previous work leveraging auxiliary information \citep{hyvarinen2019nonlinear,wang2020cost, lachapelle2021disentanglement}, we introduce distributional changes that further guarantee component-wise identifiability of latent variables.
Specifically, we assume that the change stems from an auxiliary variable $U$, which could be the domain index or time steps. In line with the approach of achieving identifiability through sparsity and most prior work on component-wise identifiability (e.g., nonlinear ICA), we assume that the latent variables are conditionally independent given $U$, i.e., $p(Z) = \prod_{i=1}^{n} p(Z_i | U)$. The identifiability is as follows.

\begin{restatable}{theorem}{change}\label{thm:change}
    Consider two models $\theta = (f,p_Z,p_{\epsilon})$ and $\hat{\theta} = (\hat{f},p_{\hat{Z}},p_{\hat{\epsilon}})$ following the process in Section \ref{sec:prelim}. In addition to the assumptions in Theorem \ref{thm:nf_suff}, suppose the following assumptions hold:
    \begin{enumerate}[label=\roman*.,ref=\roman*]
      \item The density $p_Z$ is positive and smooth. 
      \label{assum:density}
      \item (\underline{Distributional Variability}) There exist $2n + 1$ values of $U$, i.e., $U^{(i)}$ with $i \in \{0,1,\ldots,2n\}$, s.t. the $2 n$ vectors $\mathbf{w}(Z, U^{(i)}) - \mathbf{w}(Z, U^{(0)})$ with $i \in \{1,\ldots,2n\}$ are linearly independent, where vector $\mathbf{w}(Z, U)$ is defined as follows:
    \begin{equation*}
        \mathbf{w}(Z, U^{(i)}) = \left(\mathbf{v}(Z, U^{(i)}), \mathbf{v}^\prime(Z, U^{(i)})\right),
    \end{equation*}
    where
    \scriptsize
    \begin{equation*}
        \begin{aligned}
            \mathbf{v}(Z, U^{(i)}) = \Big(&\frac{\partial \log p (z_1 | U^{(i)})}{\partial z_1},\cdots,
        \frac{\partial \log p (z_{n} | U^{(i)})}{\partial z_{n}}\Big),\\
            \mathbf{v}^\prime(Z, U^{(i)}) = \Big(&\frac{{\partial}^2 \log p (z_1 | U^{(i)})}{(\partial z_1)^2},\cdots,\frac{{\partial}^2 \log p (z_{n} | U^{(i)})}{(\partial z_{n})^2}\Big).
        \end{aligned}
    \end{equation*} 
    \normalsize
    \label{assum:change}
    \vspace{-1em}
    \end{enumerate}
    Then there exists a component-wise invertible function $h$ and a permutation $\pi$ such that, $\forall i \in \{1, \ldots, n\}$,
    \begin{align*}
        p(x;\theta) = p(x;\hat{\theta}) \implies \hat{z}_i = h_i(\pi(z_i)).
    \end{align*}
\end{restatable}

The proof is in Appendix \ref{sec:proof:thm_change}. As discussed, the distributional variability (Assumption \ref{assum:change} in Thm. \ref{thm:change}) has been widely leveraged in the literature of identifiable latent variable models. Intuitively, it indicates that the auxiliary variable $U$ should have a sufficiently diverse impact on latent variables, which is usually satisfied unless the changing mechanism is almost invariant. For instance, the assumption could imply that the latent variables should evolve over time rather than remain constant. This could manifest as gradual changes in economic indicators, shifts in user behavior across different time periods, or evolving trends in datasets collected over time. Such temporal variation ensures that the underlying structure of the latent variables is exposed, facilitating their identification through changes in their distribution. 

Therefore, we conclude that under conditions of fundamentally different forms of diversity—whether structural or distributional—all latent variables can be identified component-wise, enabling the complete disentanglement of the latent generative factors. Different from existing theories, our framework is based on one of the most general settings, where we consider \textit{nonparametric}, \textit{noninvertible} generating function, in the presence of \textit{general noise}.

% Together with the injectivity of the mapping between $Z$ and $X$, we can reinterpret the identifiability result as follows.

\vspace{-0.5em}
\section{Estimation}
\label{sec:estimation}
\vspace{-0.5em}
In this section, we propose two methods for estimating the unobserved variable using observational data. The first method utilizes the KL divergence between two constructed distributions, employing a kernel-based density estimator, and is efficient for univariate unobservables. The second method employs a regularized autoencoder, designed to handle multivariate cases in the general scenarios.

% \vspace{-1em}
\vspace{-0.3em}
\subsection{Divergence-based Estimator}
\label{sec:estimation_div}
\vspace{-0.5em}

The discussion above implies that we can measure the dissimilarity between a general joint distribution
$p_{X_1, X_2,...,X_m,Z}$ 
and a distribution satisfying conditional independence $p_{ci}= p_{X_1|Z} p_{X_2|Z}...p_{X_m|Z}p_{Z}
$
in order to search for latent draws $Z_i$. One of the choices is the Kullback–Leibler divergence 
$$D_{KL} \left (p(x) || p_{ci}(x) \right ) = \int p(x) \ln  \left ( \frac{p(x)}{p_{ci}(x)} \right) dx.$$
It is worth noting that our theory requires as few as three conditionally independent groups, and the estimation methods can be easily modified to incorporate this if the grouping is known as a prior. We build a divergence-based estimator, called Generative Element Extraction Networks (GEEN), to generate the latent realizations of $Z_i$ satisfying the conditional independence. Let $\vec{V}$ stand for the vector of draws of variable $V$ in the sample, i.e., 
$\vec{Z}=(Z^{(1)},Z^{(2)},...,Z^{(N)})^T$ and 
$\vec{X}_j=(X^{(1)}_j,X^{(2)}_j,...,X^{(N)}_j)^T$. 
We generate $\vec{\hat{Z}}$ as follows: 
\begin{equation*} 
\vec{\hat{Z}}=G(\vec{X}_1,\vec{X}_2,...,\vec{X}_m).
\end{equation*}
with $\vec{\hat{Z}}=(\hat{Z}^{(1)},\hat{Z}^{(2)},...,\hat{Z}^{(N)})^T$.
The neural network $G$ is trained to minimize the divergence 
\begin{equation*} \label{equ loss}
\underset{G}{\min} \, D \left (\hat{p}\, , \, \hat{p}_{ci} \right) \quad s.t.  \int x\hat{p}_{X_1|\hat{Z}} (x|z)dx = z  
\end{equation*}
with 
\begin{align*}
    \hat{p}&=\hat{p}_{X_1, X_2,...,X_m,\hat{Z}}, \\
    \hat{p}_{ci}&=\hat{p}_{X_1|\hat{Z}} \hat{p}_{X_2|\hat{Z}}...\hat{p}_{X_m|\hat{Z}}\hat{p}_{\hat{Z}},
\end{align*}
where $\hat{p}$ are empirical distribution functions based on sample $(\vec{X}_1,\vec{X}_2,...,\vec{X}_m, \vec{\hat{Z}})$. 

Notice that $G$ enters the loss function through  $\vec{\hat{Z}}=(\hat{Z}^{(1)},\hat{Z}^{(2)},...,\hat{Z}^{(N)})^T$ in density estimators. To be specific, we can have a kernel density estimator
\begin{align*}
    &\hat{p}(x_1,\dots,x_k,\hat{z}) \\
    = &\frac{1}{m} \sum_{i=1}^m K_{h^*}\bigl(\hat{z}-\hat{Z}^{(i)}\bigr) \prod_{j=1}^k K_{h_j}\bigl(x_j - X^{(i)}_j\bigr),
\end{align*}
where
\begin{equation*}
    K_h(u) = \frac{1}{h} K\left(\frac{u}{h}\right),
\end{equation*}
and the conditional density estimator as
\begin{equation*}
    \hat{p}_{X_j|\hat{Z}}(x|\hat{z}) = \frac{\sum_{i=1}^m K_{h_j}\bigl(x-X_j^{(i)}\bigr) \, K_{h^*}\bigl(\hat{z}-\hat{Z}^{(i)}\bigr)}
    {\sum_{i=1}^m K_{h^*}\bigl(\hat{z}-\hat{Z}^{(i)}\bigr)},
\end{equation*}
where $h$ stands for bandwidths, $N$ is the total sampled observations, $m$ is the number of points in each observation and $k$ is the number of features. 

% In the loss function defined in equation (\ref{equ loss}), it requires more than one data point to estimate the kernel density function. As a result, unlike other use cases that one training point is enough to calculate its corresponding loss, we need to sample $m$ ($>1$) points as one observation to calculate its loss. For example, to build the training sample we sample with replacement $m$ points from the entire training data points and repeat $N$ times, and we end up with $N$ observations in our training sample. The same practice is followed to construct our validation and test samples. The kernel function $K(\cdot)$ can simply be the standard normal density function. For the bandwidth, we adopt the so-called Silverman's rule, i.e., $h^j = w \sigma^j N^{-1/5}$ where $ \sigma^j$ is the standard error of $X^j$, and $w$ is the window size that is determined by hyper parameters tuning. Similarly, we may take $h^* = w \sigma^* N^{-1/5}$, where $ \sigma^*$ is the standard error of $Z$. 

\subsection{Regularized Autoencoder}
\label{sec:estimation_vae}

Since kernel density estimation suffers from the curse of dimensionality, in which the number of computations required increases exponentially with the number of dimensions, we propose a regularized autoencoder-based estimator to deal with multivariate cases. Unlike traditional losses, we need to incorporate the conditional independence constraints. The $i$-th observed variable $X_i$ is generated as $X_i = f_i(Z, \epsilon_i)$. The log-likelihood can be transformed as follows:
\begin{align*}
    &\log \hat{p}(X | \hat{Z}) 
    = \sum_i \log \hat{p}(X_i | \hat{Z}) \\
    = &\sum_i \log \left( \frac{\hat{p}(\hat{\epsilon}_i)}{\left|\frac{\partial \hat{f}_i}{\partial \hat{\epsilon}_i}\right|} \right) = \sum_i \left( \log \hat{p}(\hat{\epsilon}^i) - \log \left| \frac{\partial \hat{f}^i}{\partial \hat{\epsilon}^i} \right| \right).
\end{align*}
Thus, the loss of our regularized autoencoder is defined as:
\begin{align*}
    \mathcal{L}_{\text{RAE}} = -\log \hat{p}(X | \hat{Z}) + D_{KL} \left( [\hat{p}(\hat{Z}), \hat{p}(\hat{\epsilon})] \, \| \, \mathcal{N}(0, \mathbb{I}) \right),
\end{align*}
where we use KL divergence to enforce the independence among components in $\hat{X}$ and $\hat{\epsilon}$, and $\mathbb{I}$ is an identity matrix. 

\section{Experiments}

In this section, we conduct experiments on both synthetic and real-world datasets to verify our claims. Additional details are included in Appendix \ref{appx:experiments}.

\vspace{-0.5em}
\subsection{Simulations}
\vspace{-0.2em}

\textbf{Basis validation.} \
We first conduct experiments on the basic setting to evaluate the identification from observations. The samples are generated as follows:
\begin{equation}
    X^{(i)}_j = f_j(Z^{i}) + \epsilon^{(i)}_j
\end{equation}
for $j=1,2,...,k$ and $i=1,2,...,N$.
Without loss of generality, we normalize
$
f_1(x)= x
$
and 
$ \mathbb{E}[\epsilon_1 | Z]=0. $
We pick distributions for $(\epsilon_1,...,\epsilon_k, X^*)$ and functions $(f_2,...,f_k)$ to generate a sample $(X_1,...,X_k, Z )$. In this setting, we focus on directly validating the proposed theory, and thus we start with a single latent variable. Thus, we use the divergence-based estimator GEEN (Section \ref{sec:estimation_div}) and train $G$ using the observed sample $(\vec{X}_1,\vec{X}_2,...,\vec{X}_k)$ to generate $(\vec{X}_1,\vec{X}_2,...,\vec{X}_k, \hat{Z})$. That is $\vec{\hat{Z}} = G(\vec{X}_1,\vec{X}_2,...,\vec{X}_k)$. 
% We sample $Z$ from $\mathcal{N}(0,4)$ and the $\epsilon$ from $\mathcal{N}(0,1)$ (\textit{Baseline}) or $\mathcal{N}(0,1)$
For the baseline case, we consider the following generating process: 
\begin{equation*}
\begin{aligned}
    k = 4&, \quad  \epsilon_1 \sim \mathcal{N}(0, 1), \\
    f_1(z) = z&, \quad  \epsilon_2 \sim Beta(2, 2) - \frac{1}{2}, \\
    f_2(z) = \frac{1}{1+e^z}&, \quad  \epsilon_3 \sim Laplace(0, 1), \\
    f_3(z) = z^2&, \quad  \epsilon_4 \sim Bernoulli\left(\frac{1}{2}\right),\\
    X_4 = \Phi(Z/3) \cdot &(-1)^{I(\epsilon_4 > 0.5)}, \quad  Z \sim \mathcal{N}(0, 4).
\end{aligned}
\end{equation*}

We sample 8000 points as training points from the above distributions for $Z$, $\epsilon_1$, $\epsilon_2$, $\epsilon_3$ and $\epsilon_4$. Then we sample another 1000 points for validation points and 1000 points for test points. We draw 500 points from the training points with replacement 8000 times to build our training set and 1000 times from the validation/test points to build our validation/test set. In the second experiment, we let the error terms correlate with $Z$ while keeping the rest of the setup the same as the baseline. Specifically, we use:
\begin{equation*}
\begin{split}
\epsilon_1 & =  \mathcal{N}(0, \frac{1}{4} z^2), \\
\end{split}
\quad
\begin{split}
\epsilon_3 & =  Laplace(0, \frac{1}{2}|z|). \\
\end{split}
\end{equation*}
In the third experiment, we double the variance of the error terms while keeping the rest setup the same as the baseline:
\begin{equation*}
\epsilon_1 \sim \mathcal{N}(0, 4), \epsilon_2 \sim Beta(2, 4) - \frac{1}{3}, \epsilon_3 \sim Laplace(0, 2).
\end{equation*}

\begin{table}[t]
\centering
\caption[Basis validation for continuous data]{Basis Validation for Continuous Data}
\label{table:simmulation_summary}
%\begin{center}
%\captionsetup{justification=centering}
\resizebox{0.46\textwidth}{!} {
{
\begin{threeparttable}
\begin{tabular}{l*{4}{c}}
 \toprule
 Simulation Name &\multicolumn{3}{c}{corr($\vec{Z}$, $\vec{\hat{Z}}$)}
 & corr($\vec{Z}$, $\vec{X_1}$)\\
  & min & median & max  & \\
 \cmidrule{2-4}
 Baseline  & 0.97  & 0.98 & 0.98 & 0.89  \\

Linear Error  & 0.93  & 0.96 & 0.97 & 0.89  \\

Double Error  & 0.80  & 0.89 & 0.91 & 0.70  \\
\bottomrule
\end{tabular}
\end{threeparttable}
}
}
\vspace{-1em}
%\end{center}
\end{table}

Table \ref{table:simmulation_summary} demonstrates the min, median and max correlations of $\vec{Z}$ and $\vec{\hat{Z}}$ in the test sample for the three experiments after running each one 25 times. It shows that the estimation is robust with randomly picked initial values of the parameters and provides a better measurement of $Z$ than $X_1$. The correlation between $Z$ and the generated $Z$ is well above 0.9 for the baseline and the linear error case and remains strong when the variance is doubled in the third experiment.

In addition to generating continuous $Z$, we will also demonstrate how our method performs with discrete $Z$.  We have the same set-up as the continuous examples except that now we sample $Z$ from the binomial distribution. Similar to the continuous case, we also have three different settings for discrete data, i.e., baseline, linear error, and double error. Details of the data generating process are included in Appendix \ref{appx:experiments}.

When $Z$ are sampled from discrete random variables, the task to identify $Z$ is similar to an unsupervised clustering task. Therefore, we also compare our method with k-means. In order to facilitate direct comparison, when running the k-means algorithm, we set the number of clusters equal to $11$ and randomly pick one point from each cluster (Cluster k is the set of points with $Z=k$) as initial points. With this setup, k-means is actually put in an advantageous position since $Z$ is completely unkown to our estimator, but k-means is provided with limited information of the clusters (e.g. the number of clusters and initial point from each cluster). As shown in Table \ref{tabel:discrete-compare}, in all three cases, our estimator has better performance than k-means when measuring the correlation between $Z$ and $\hat{X}_*$, especially for linear error and double error cases.

\begin{table}[t]
\centering
\caption{Comparison of corr($\vec{Z}$, $\vec{\hat{Z}}$) between our estimator and k-means for discrete data.}
\begin{tabular}{lll}
\hline
Simulation Name & k-means & GEEN \\ \hline
Baseline        & 0.98   & 0.99 \\
Linear Error    & 0.96   & 0.97 \\
Double Error    & 0.97   & 0.98 \\ \hline
\end{tabular}
\vspace{-1em}
\label{tabel:discrete-compare}
\end{table}

\textbf{Generalized validation.} \
\looseness=-1
In the previous experiments, we have carefully validated our theoretical claims in various settings. Now we would like to explore the estimation of latent variables in the general settings, where there are multiple latent variables. Thus, we conduct experiments on the regularized-autoencoder-based estimator (Section \ref{sec:estimation_vae}). For each latent variable ${Z}_i \sim \mathcal{N}(0,4)$, we have three observed variables generated from ${Z}_i$ by a nonlinear transformation together with a noise $\epsilon_i \sim \mathcal{N}(0,\sigma^2)$, satisfying the structural sparsity condition. Following previous work \citep{hyvarinen2024identifiability}, we use mean correlation coefficient (MCC) between the true latent variables and the estimated ones as the evaluation metric. Loosely speaking, a higher MCC indicates a more disentangled recovered latent representation, which quantifies the identification quality of multiple latent variables. 
\begin{figure}
    \centering
    \includegraphics[width=0.9\linewidth]{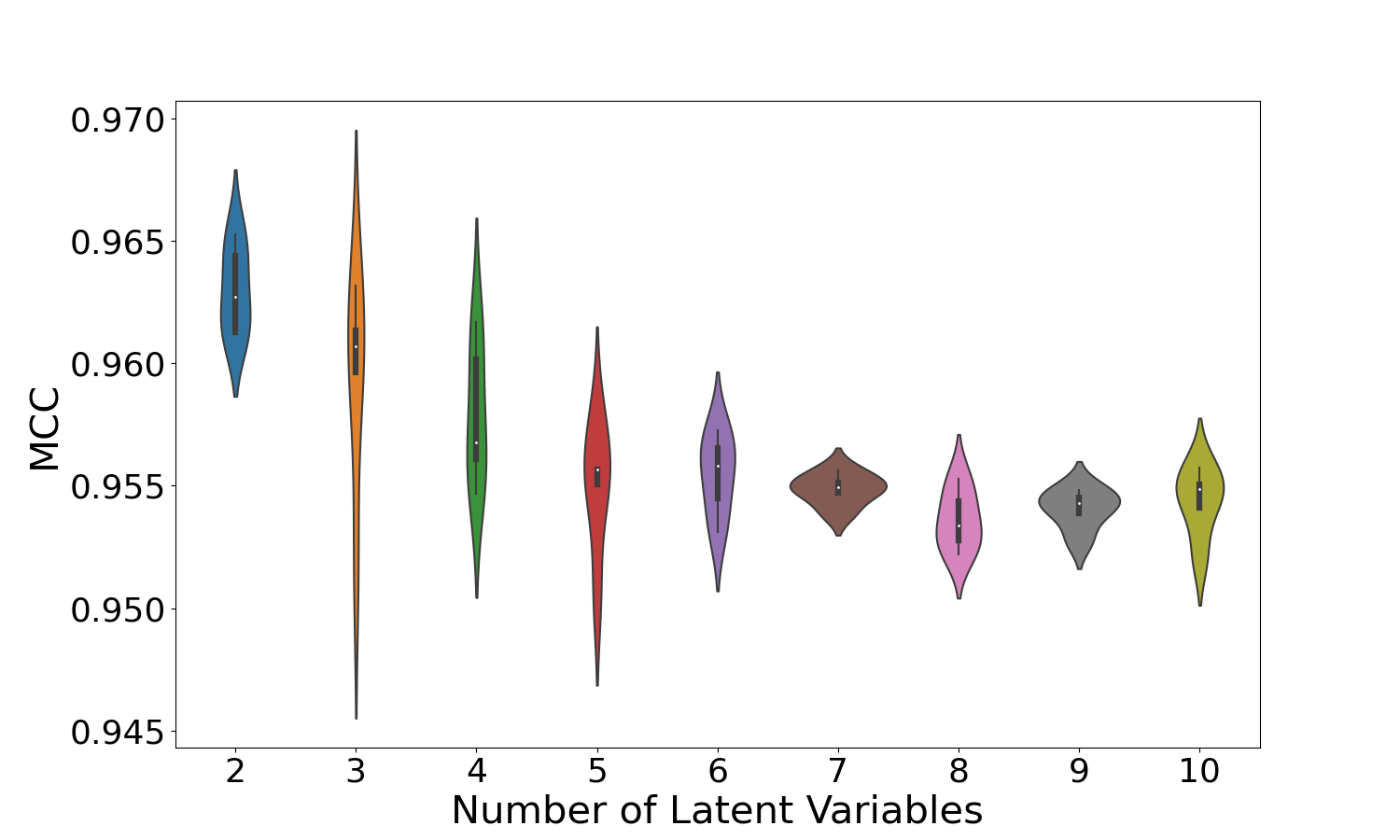}
    \caption{Results w.r.t. different numbers of latent variables.}
    \vspace{-0.5em}
    \label{fig:latent_num}
\end{figure}

\begin{figure}
    \centering
    \vspace{-1em}
\includegraphics[width=0.9\linewidth]{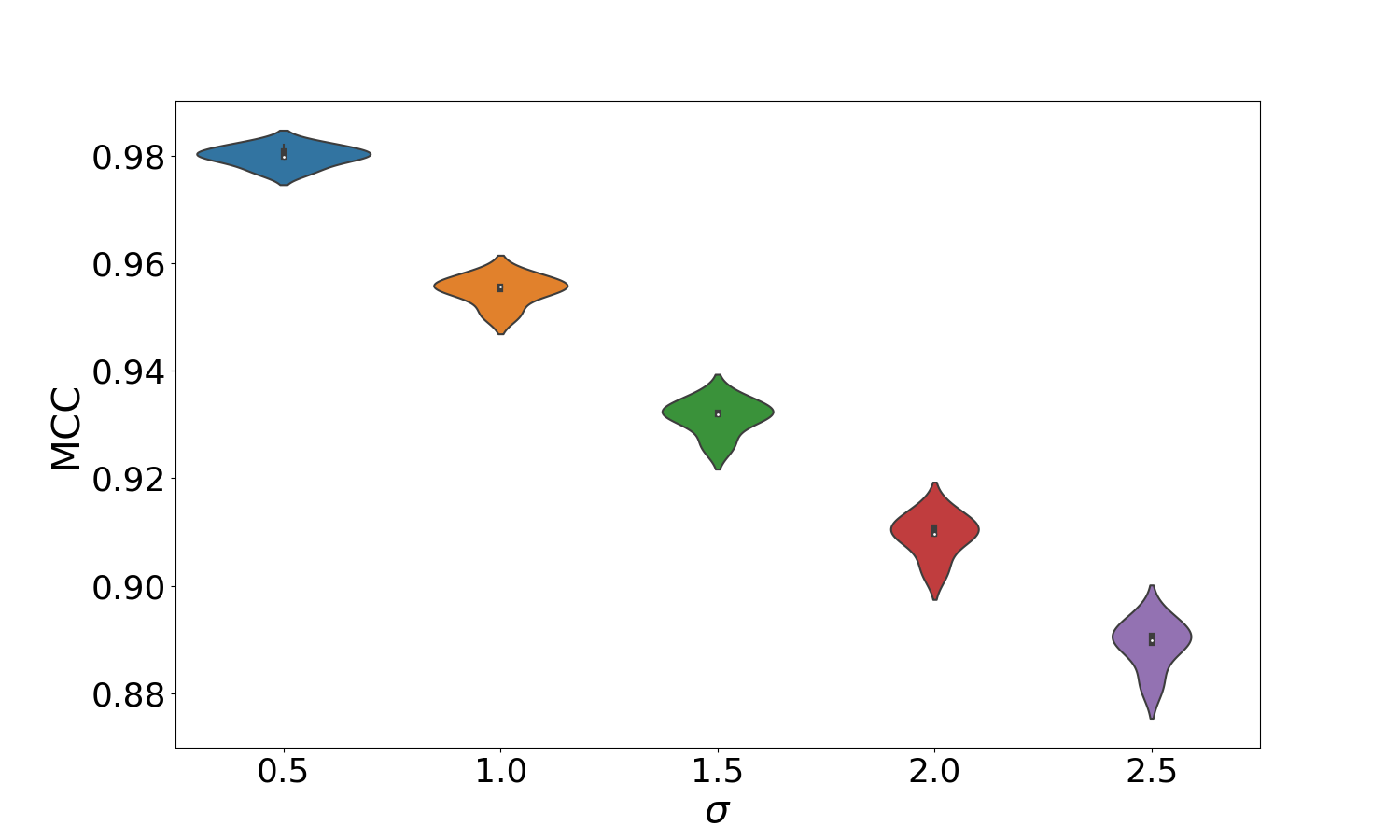}
    \caption{Results w.r.t. different standard deviations ($\sigma$) of the noise.}
    \vspace{-1.7em}
    \label{fig:noise_level}
\end{figure}

We begin by conducting experiments with a noise variance of one, i.e., $\epsilon_i \sim \mathcal{N}(0,1)$, while varying the number of latent variables from 2 to 10. The results, presented in Figure \ref{fig:latent_num}, show that our estimator consistently achieves MCC values close to one across datasets of different dimensions, demonstrating its effectiveness in general multivariate settings.

We further evaluate the estimator under varying noise levels by adjusting the standard deviation $\sigma$ in $\epsilon_i \sim \mathcal{N}(0, \sigma^2)$. The results, shown in Figure \ref{fig:noise_level}, reveal a slight performance decline as the noise level increases. However, even with a substantial noise variance, the method continues to achieve great identification results, highlighting its robustness and effectiveness.

\subsection{Refining Official GDP Measurements}

One of the important applications of our methodology is to reduce measurement errors. In this case, true values are unobservables $Z$. $X_1$ is a direct measure of $Z$ with the expected measurement error $\delta$ as zero. $X_j$ ($j\neq1$) are indirect/direct measures of $Z$ with unknown function forms of $Z$. Their error terms can be flexible and do not necessarily have zero means. 

In this section, we conduct real-world experiments to study the implications of our method in practice. We apply the divergence-based estimator (GEEN) to refine GDP data using official GDP data ($X_1$) and alternative measures of economic activity, including satellite-recorded nighttime light \citep{hu2022illuminating} and Google Search Volume \citep{woloszko2021tracking} as $X_2$ and $X_3$. In this experiment, true GDP ($Z$) are completely unknown. We demonstrate how our method can help reduce measurement errors from official GDP data. 

Our sample consists of all the developing countries that have quarterly GDP data. We focus on developing countries because nighttime light data are more appropriate for tracking economic activity in those countries \citep{hu2022illuminating, beyer2022measuring}. To account for time trends common to all countries, we remove time effects from official GDP growth rates with a fixed effect model when training and later add back the time effects to reconstruct our generated true GDP growth rates when comparing our model's performance with official data. We separate our sample into training and validation subsets, and run training 100 times and select the best model with the lowest loss in the validation sample to minimize the impact of initialization. We do not have the testing dataset, since in this case true values $Z$ are completely unknown and the model just learns how to generate $Z$ that can minimize the distance between the two probability densities in equation (\ref{equ 200}). Therefore, conventional testing method is not applicable here. Instead, we compare our generated GDP growth rates with official data from the macroeconomic viewpoint, which is crucial to reveal systematic differences between official data and true underlying GDP growth data.   

\begin{figure}[t]
    \centering
    \includegraphics[width = 0.47\columnwidth]{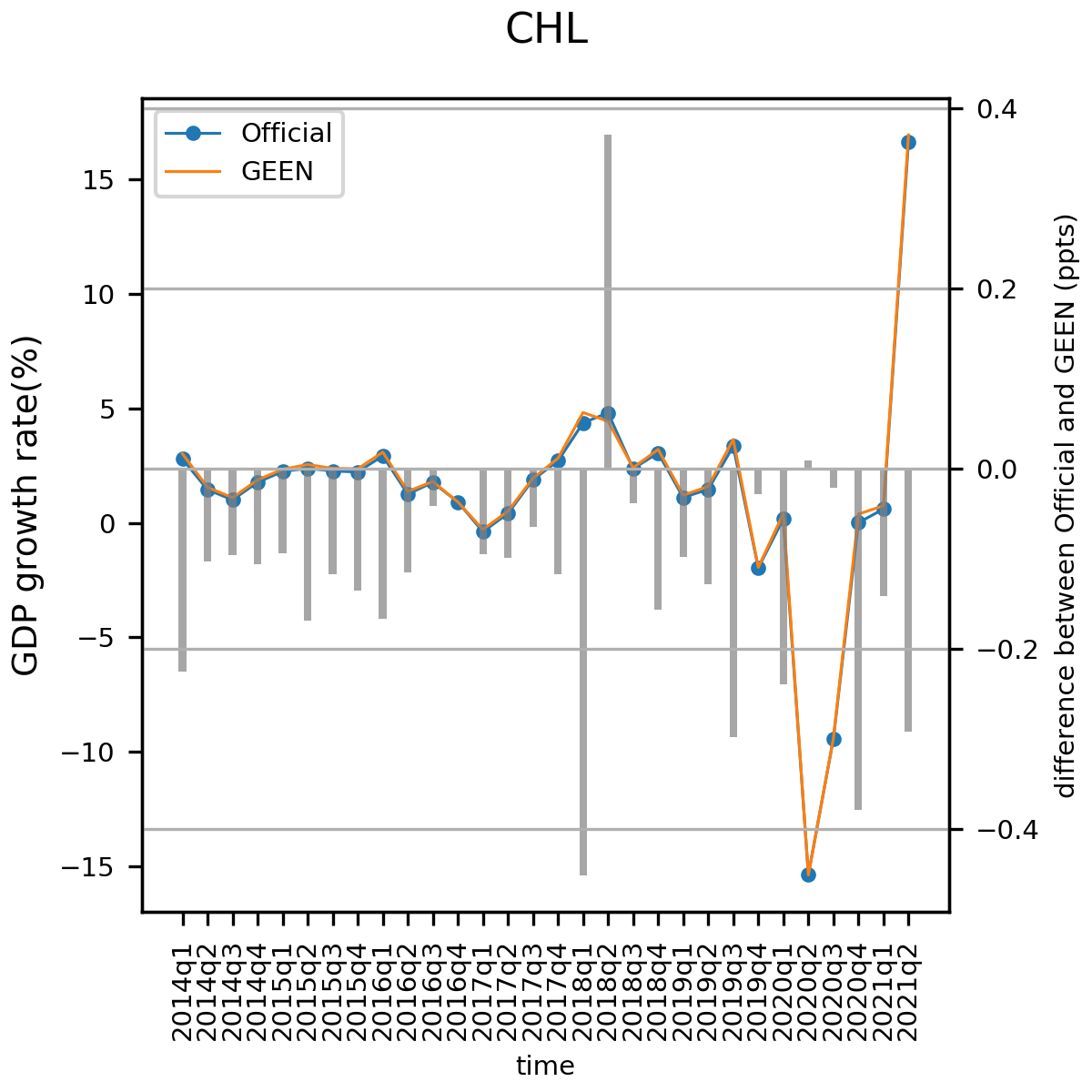}\includegraphics[width = 0.47\columnwidth]{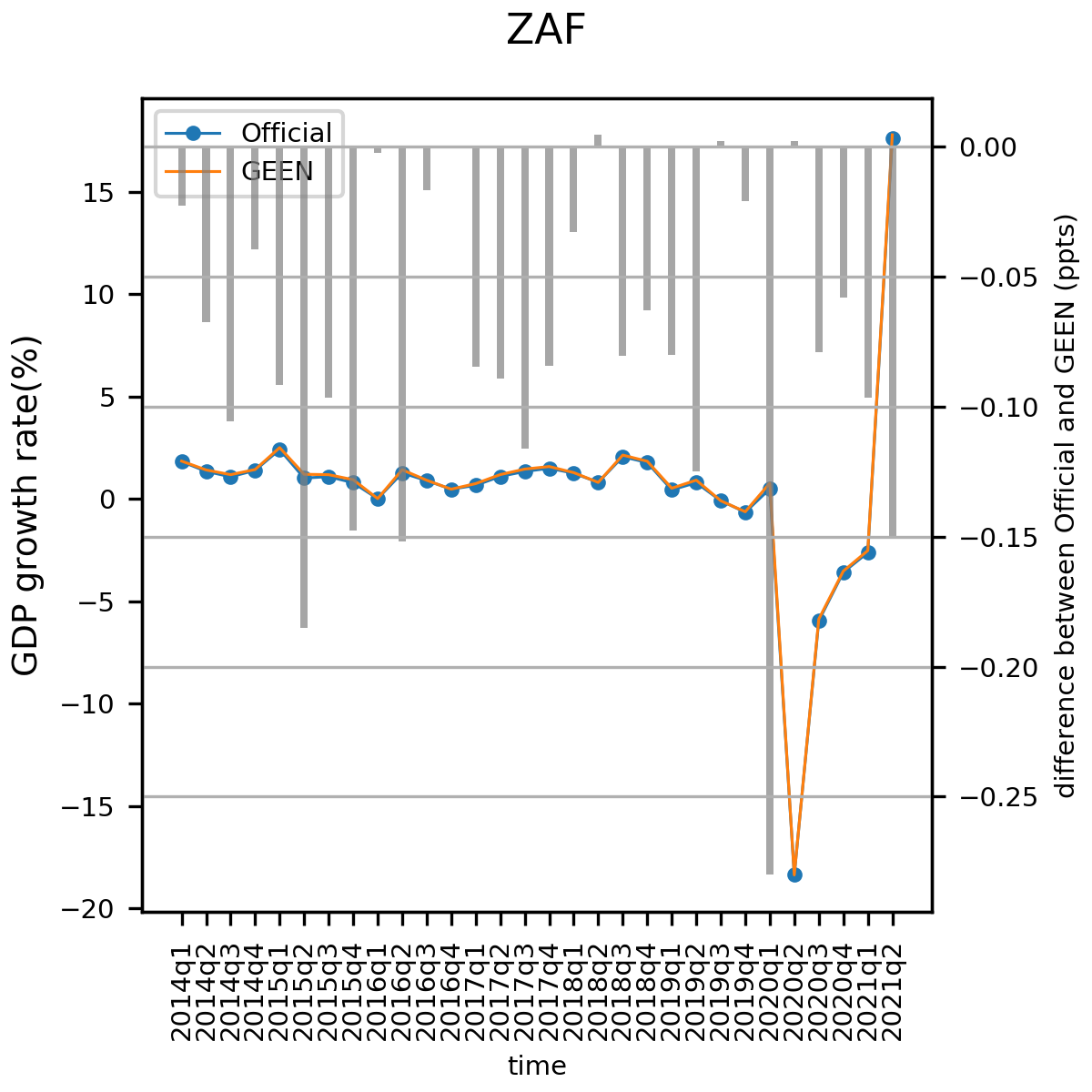}
    \includegraphics[width = 0.47\columnwidth]{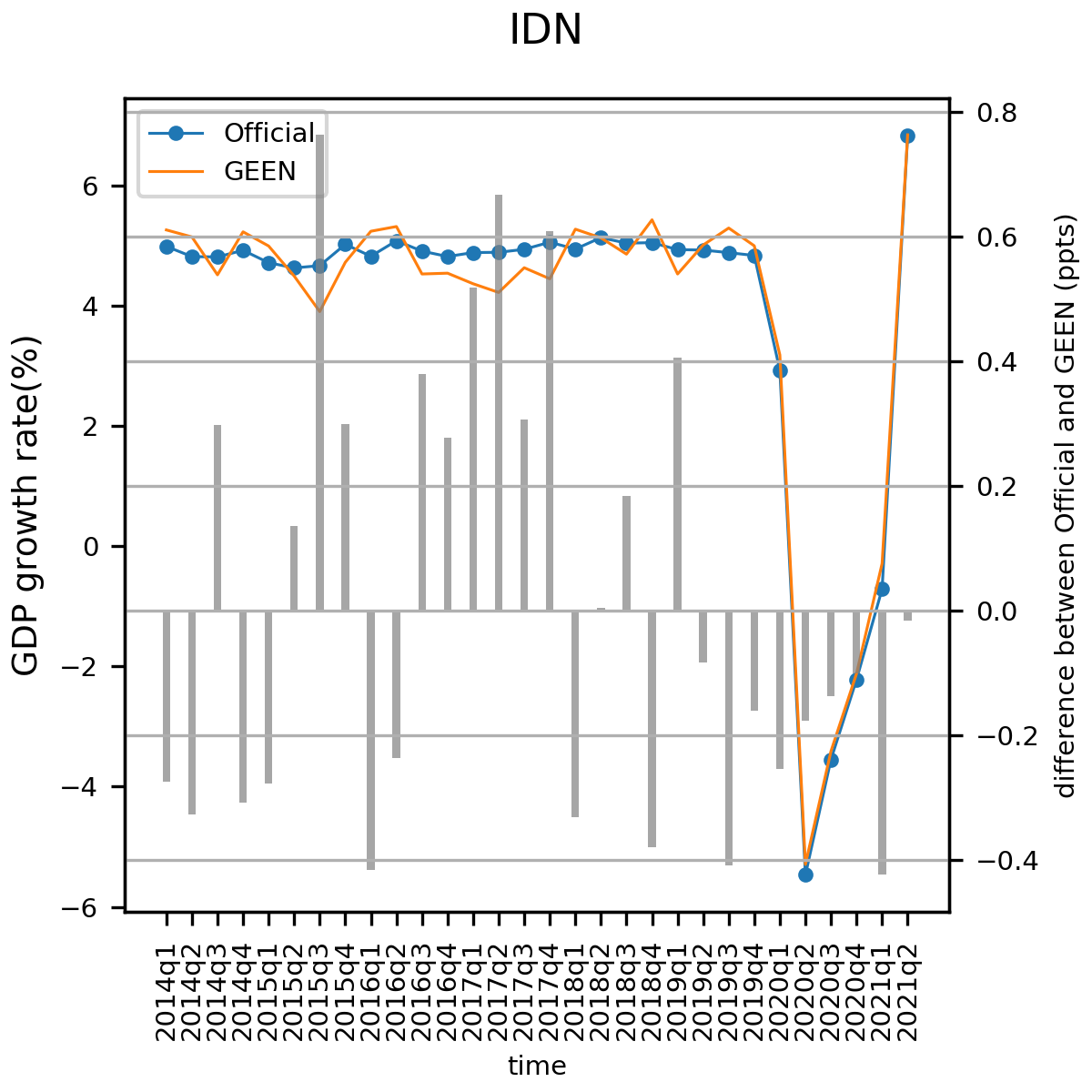}\includegraphics[width = 0.47\columnwidth]{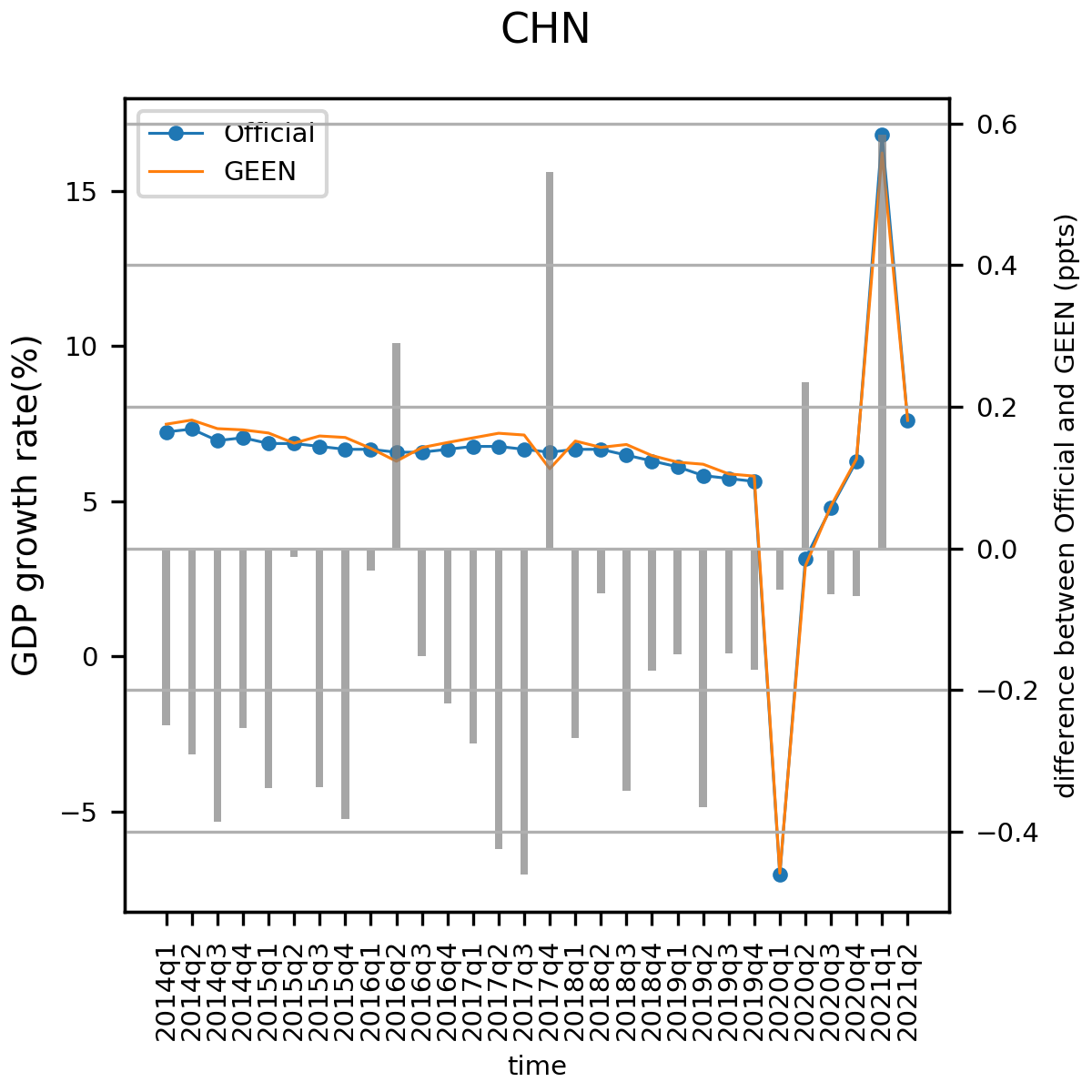}
     \includegraphics[width = 0.47\columnwidth]{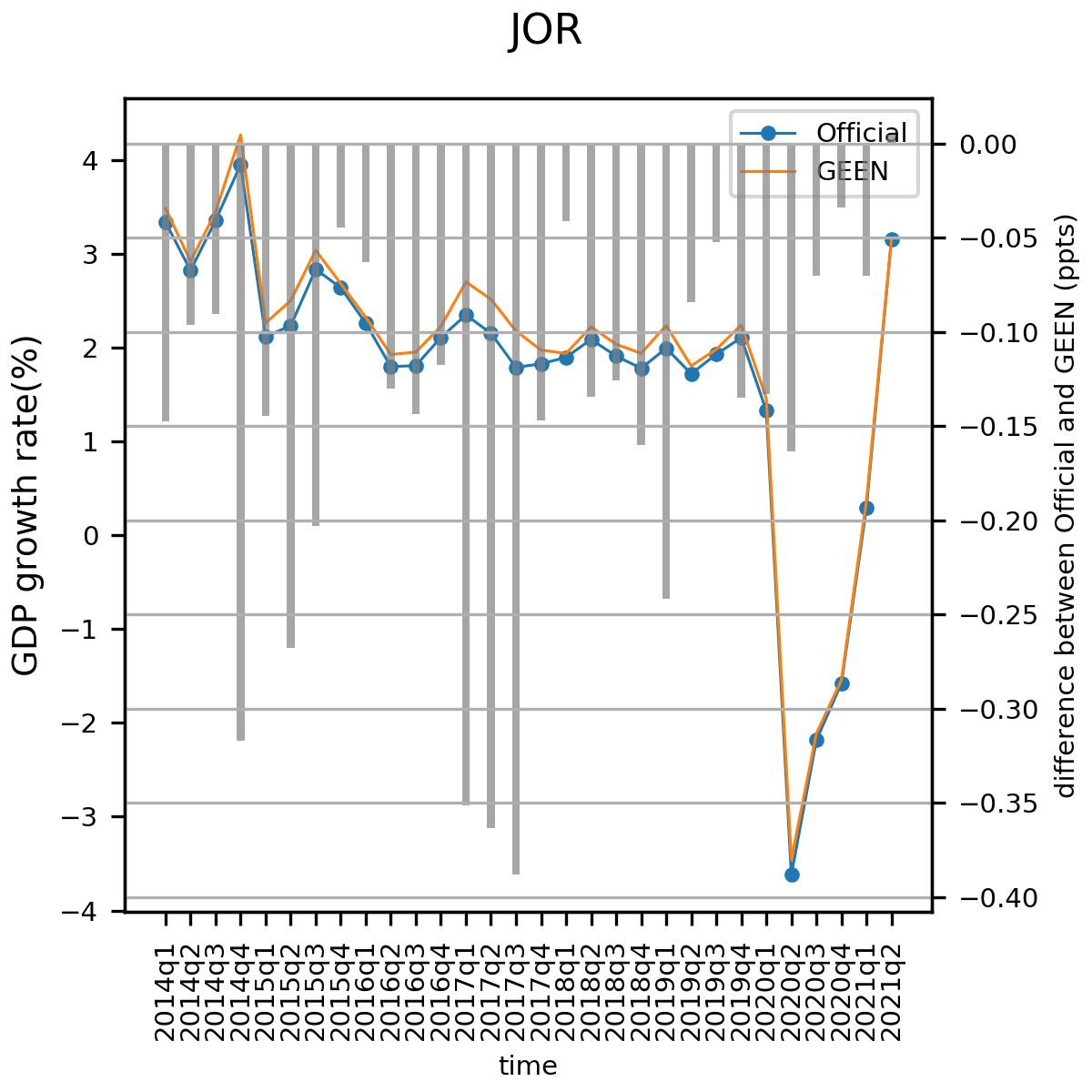}\includegraphics[width = 0.47\columnwidth]{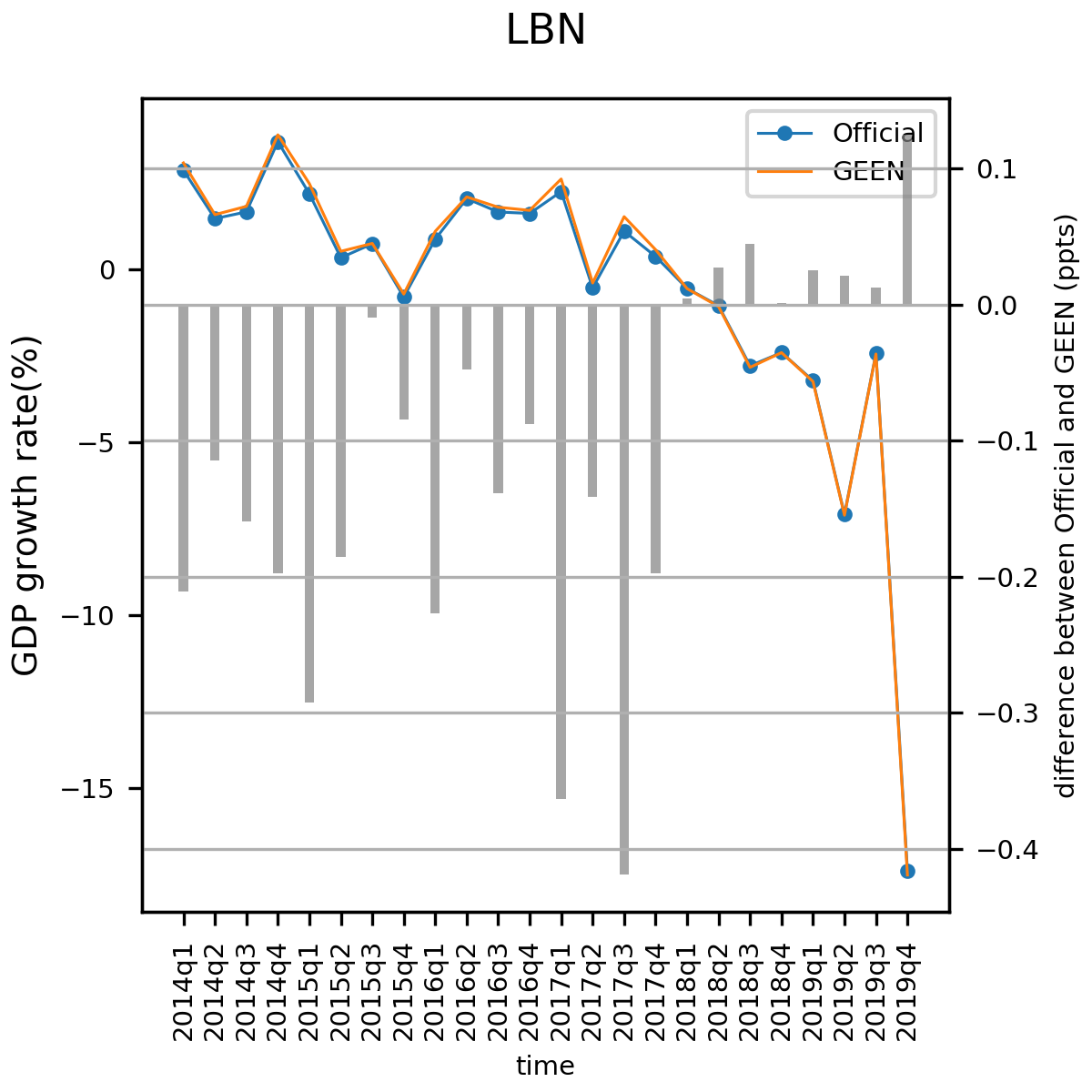}
    \caption{Country examples of official and refined GDP growth.}
    \label{fig: GDP results}
    \vspace{-1em}
\end{figure}

In Figure \ref{fig: GDP results}, the left axis is GDP growth rate in percentage points (ppts) and the right axis marks the difference between the official GDP and our generated underlying GDP growth rates (Official - GEEN as shown in the plot). Figure \ref{fig: GDP results} shows that refined GDP data reveal important patterns in official GDP data and are useful in a number of aspects. First, most countries' official GDP growth data align well with our refined estimates. For example, both Chile and South Africa have differences within 0.15 percentage points despite volatile economic growth. It suggests that GEEN would not contradict official GDP data when they are realatively accurate and could possibly improve upon them. 

Second, some countries, such as China and Indonesia, have excessively smooth official GDP data compared to our refined estimates. Such excess smoothness might mask underlying dynamics and volatility of economic activity (for countries like Indonesia and China, an adjustment of 0.5 percentage points in GDP is considered significant). Estimates of underlying economic growth could therefore enrich policymakers' understanding of the state of macroeconomy, including output gap and inflationary pressures, and inform efficient policy making. 

Third, the official GDP growth data of some economies systematically diverge from our refined estimates. For instance, when Lebanon's economy contracted after 2017, the official data consistently overstated its performance, whereas Jordan's official data tended to understate economic growth. A likely explanation is the presence of informal sectors not captured by official statistics. Identifying these discrepancies is a crucial first step in investigating their underlying causes, whether they relate to the statistical agency's capacity, the recording of informal economic activity, or factors within the political economy. 

\vspace{-0.5em}
\section{Conclusion}
\vspace{-0.5em}

We have established a comprehensive framework for one of the most general settings in latent variable identification. Specifically, we prove the identifiability of latent variables in nonparametric models with nonlinear, noninvertible generating processes, even when confronted with general non-negligible noise. We show that, under standard conditions such as structural or distributional variability, latent variables in these nonlinear models can be identified up to minor ambiguities, despite the presence of complex noise. Building on the theoretical foundation, we developed estimation methods and validated them through extensive experiments on both synthetic and real-world data. The results indicate a strong correlation between the estimated and true values across various scenarios. Additionally, our analysis of real-world economic data reveals that our method can uncover more detailed insights than those provided by official reports. While our study might be limited by a lack of downstream applications across a broader range of fields, we believe our framework has the potential to transform the way researchers address latent variables in practice.

% The simulation results show that the proposed estimators work quite well and the estimated values are highly correlated with the true values in various settings. We then estimate the true GDP growth for each developing country using its official GDP growth, nightlight intensity, and Google search volume. For different types of economies, our estimates show more meaningful and insightful information than official GDP growth. We expect our method will change how researchers deal with latent variables in empirical research.

\section*{Acknowledgement}
We appreciate the anonymous reviewers for their constructive feedback. We would also like to acknowledge the support from NSF Award No. 2229881, AI Institute for Societal Decision Making (AI-SDM), the National Institutes of Health (NIH) under Contract R01HL159805, and grants from Quris AI, Florin Court Capital, and MBZUAI-WIS Joint Program.

\bibliography{bibliography}
\bibliographystyle{plainnat}

\newpage
\onecolumn
\aistatstitle{Supplement to ``Nonparametric Factor Analysis and Beyond"}
\appendix

\part{} % Start the appendix part

{\hypersetup{linkcolor=black}
\parttoc % Insert the appendix TOC
}

\newpage

\section{Summary of Notation}
\label{appx:notation}

This appendix summarizes the notation used throughout the paper for easy reference.

\subsection*{Variables}

We consider the following variables:

\begin{itemize}
    \item \emph{Observed variables}: \( X = (X_1, X_2, \ldots, X_m) \), where \( X \in \mathcal{X} \subseteq \mathbb{R}^m \).
    \item \emph{Latent variables}: \( Z = (Z_1, Z_2, \ldots, Z_n) \), with \( Z \in \mathcal{Z} \subseteq \mathbb{R}^n \).
    \item \emph{Noise variables}: \( \epsilon = (\epsilon_1, \epsilon_2, \ldots, \epsilon_e) \), where \( \epsilon \in \mathcal{E} \subseteq \mathbb{R}^e \).
\end{itemize}

\subsection*{Data-Generating Process}

The data are generated according to the function:
\begin{equation*}
    X = f(Z, \epsilon).
\end{equation*}
Note that \( Z \) and \( \epsilon \) may not be independent (it is possible that \( Z \not\!\perp\!\!\!\perp \epsilon \)), and thus the function \( f \) is generally non-invertible.

\subsection*{Probability Notation}

\begin{itemize}
    \item Random variables are denoted by capital letters (e.g., \( V \)); realizations are denoted by lowercase letters (e.g., \( v \)).
    \item The probability density function (pdf) of \( V \) is denoted by \( p_V(v) \).
    \item The conditional density of \( V \) given \( U \) is denoted by \( p_{V|U}(v|u) \).
    \item Probability distributions are denoted by \( P \).
\end{itemize}

\subsection*{Matrix Notation}

For any matrix \( S \):

\begin{itemize}
    \item \( S_{i,:} \) denotes the \( i \)-th row of \( S \).
    \item \( S_{:,j} \) denotes the \( j \)-th column of \( S \).
\end{itemize}

Given an index set \( \mathcal{I} \subseteq \{1, \ldots, a\} \times \{1, \ldots, b\} \), we define:

\begin{align*}
    \mathcal{I}_{i,:} &\coloneqq \{ j \mid (i, j) \in \mathcal{I} \}, \\
    \mathcal{I}_{:,j} &\coloneqq \{ i \mid (i, j) \in \mathcal{I} \}.
\end{align*}

\subsection*{Supports and Subspaces}

\begin{itemize}
    \item The support of a matrix \( S \in \mathbb{R}^{a \times b} \) is defined as:
    \begin{equation*}
        \operatorname{supp}(S) \coloneqq \{ (i, j) \mid S_{i,j} \ne 0 \}.
    \end{equation*}
    \item The support of a matrix-valued function \( \mathbf{S}(\Theta): \Theta \rightarrow \mathbb{R}^{a \times b} \) is defined as:
    \begin{equation*}
        \operatorname{supp}(\mathbf{S}(\Theta)) \coloneqq \{ (i, j) \mid \exists \theta \in \Theta, \ \mathbf{S}(\theta)_{i,j} \ne 0 \}.
    \end{equation*}
    \item For any subset \( \mathcal{S} \subseteq \{1, \ldots, n\} \), we define the subspace:
    \begin{equation*}
        \mathbb{R}_{\mathcal{S}}^n \coloneqq \{ s \in \mathbb{R}^n \mid s_i = 0 \text{ for all } i \notin \mathcal{S} \}.
    \end{equation*}
\end{itemize}

\subsection*{Jacobians}

\begin{itemize}
    \item The Jacobian of \( f \) with respect to \( Z \) is denoted by \( J_f \).
    \item The support of \( J_f \) is \( \mathcal{F} = \operatorname{supp}(J_f) \).
\end{itemize}

\subsection*{Sets of Matrices}

We define \( \mathcal{T} \) as the set of matrices that share the same support as \( \mathbf{T} \) in the equation:
\begin{equation*}
    J_{\hat{f}} = J_f \mathbf{T},
\end{equation*}
where \( \mathbf{T} \) is a matrix-valued function.

\subsection*{Estimated Quantities}

Estimated quantities are denoted with a hat symbol, for example:
\begin{itemize}
    \item \( \hat{Z} \) denotes an estimate of \( Z \).
    \item \( \hat{f} \) denotes an estimate of \( f \).
\end{itemize}

\subsection*{Function Spaces}

The space of bounded and integrable functions on \( \mathcal{Z} \) is defined as:
\begin{equation*}
    \mathcal{L}_{\text{bnd}}^1(\mathcal{Z}) \coloneqq \left\{ g : \int_{\mathcal{Z}} |g(z)| dz < \infty,\ \sup_{z \in \mathcal{Z}} |g(z)| < \infty \right\}.
\end{equation*}

\subsection*{Linear Operators}

The integral operator corresponding to \( p_{X_1|Z} \) is defined by:
\begin{align*}
    L_{X_1|Z} &: \mathcal{L}_{\text{bnd}}^1(\mathcal{Z}) \rightarrow \mathcal{L}_{\text{bnd}}^1(\mathcal{X}_1), \\
    (L_{X_1|Z} h)(x) &= \int_{\mathcal{Z}} p_{X_1|Z}(x|Z) \, h(Z) \, dZ.
\end{align*}

\subsection*{Vector Functions}

For vectors \( \mathbf{v}(Z, U^{(i)}) \) and \( \mathbf{v}'(Z, U^{(i)}) \), we define:
\begin{align*}
    \mathbf{v}(Z, U^{(i)}) &= \left( \frac{\partial \log p(z_1 \mid U^{(i)})}{\partial z_1}, \ldots, \frac{\partial \log p(z_n \mid U^{(i)})}{\partial z_n} \right), \\
    \mathbf{v}'(Z, U^{(i)}) &= \left( \frac{\partial^2 \log p(z_1 \mid U^{(i)})}{\partial z_1^2}, \ldots, \frac{\partial^2 \log p(z_n \mid U^{(i)})}{\partial z_n^2} \right).
\end{align*}
Then, the vector \( \mathbf{w}(Z, U^{(i)}) \) is given by:
\begin{equation*}
    \mathbf{w}(Z, U^{(i)}) = \left( \mathbf{v}(Z, U^{(i)}),\ \mathbf{v}'(Z, U^{(i)}) \right).
\end{equation*}

\section{Proofs}
\label{appx:proofs}

\subsection{Proof of Theorem \ref{thm:nf_suff}}\label{sec:proof_nf_suff}

\nfsuff*

\begin{proof}

According to Assumption \ref{assumption 3.0}, the observed variables \( X \) can be partitioned into three subsets \( \{X_A, X_B, X_C\} \), where variables in each subset are conditionally independent of those in the other subsets given \( Z \). Let us start with
\begin{equation}
    p_{X_C X_A | X_B}(x_C, x_A | x_B) = \int p_{X_C X_A Z | X_B}(x_C, x_A, z | x_B) \, dz.
\end{equation}
Applying the chain rule of conditional probability, we decompose the joint density in the integral:
\begin{equation}
    p_{X_C X_A Z | X_B}(x_C, x_A, z | x_B) = p_{X_C | X_A Z X_B}(x_C | x_A, z, x_B) p_{X_A Z | X_B}(x_A, z | x_B).
\end{equation}
Substituting this into the previous equation, we obtain:
\begin{equation}
    p_{X_C X_A | X_B}(x_C, x_A | x_B) = \int p_{X_C | X_A Z X_B}(x_C | x_A, z, x_B) p_{X_A Z | X_B}(x_A, z | x_B) \, dz.
\end{equation}

By the conditional independence structure, we have \( X_C \perp X_B \mid (X_A, Z) \), which simplifies the first conditional density:
\begin{equation}
    p_{X_C | X_A Z X_B}(x_C | x_A, z, x_B) = p_{X_C | X_A Z}(x_C | x_A, z).
\end{equation}
Thus, we substitute this into the previous expression:
\begin{equation}
    p_{X_C X_A | X_B}(x_C, x_A | x_B) = \int p_{X_C | X_A Z}(x_C | x_A, z) p_{X_A Z | X_B}(x_A, z | x_B) \, dz.
\end{equation}

Next, we apply the decomposition of \( p_{X_A Z | X_B} \). Since \( X_A \perp X_B \mid Z \), we have:
\begin{equation}
    p_{X_A Z | X_B}(x_A, z | x_B) = p_{X_A | Z X_B}(x_A | z, x_B) p_{Z | X_B}(z | x_B).
\end{equation}
Substituting this into the integral:
\begin{equation}
    p_{X_C X_A | X_B}(x_C, x_A | x_B) = \int p_{X_C | X_A Z}(x_C | x_A, z) p_{X_A | Z X_B}(x_A | z, x_B) p_{Z | X_B}(z | x_B) \, dz.
\end{equation}

Under Assumption \ref{assumption 3.0}, \( X_A \perp X_B \mid Z \) further simplifies the second conditional probability:
\begin{equation}
    p_{X_A | Z, X_B}(x_A | z, x_B) = p_{X_A | Z}(x_A | z).
\end{equation}
Thus, substituting this into the previous equation, we obtain:
\begin{equation}
    p_{X_C X_A | X_B}(x_C, x_A | x_B) = \int p_{X_C | X_A Z}(x_C | x_A, z) p_{X_A | Z}(x_A | z) p_{Z | X_B}(z | x_B) \, dz.
\end{equation}

Finally, Assumption \ref{assumption 3.0} states that \( X_C \perp X_A \mid Z \), which implies:
\begin{equation}
    p_{X_C | X_A Z}(x_C | x_A, z) = p_{X_C | Z}(x_C | z).
\end{equation}
Substituting this, we obtain the form:
\begin{equation}
    p_{X_C X_A | X_B}(x_C, x_A | x_B) = \int p_{X_C | Z}(x_C | z) p_{X_A | Z}(x_A | z) p_{Z | X_B}(z | x_B) \, dz.
\end{equation}

Let \( S \) and \( V \) be random variables with supports \( \mathcal{S} \) and \( \mathcal{V} \), respectively. A kernel operator \( K_{V|S} \) is defined as a mapping from a function \( f' \) in a function space \( \mathcal{F}(\mathcal{S}) \) onto a function \( K_{S|V} f' \) in \( \mathcal{F}(\mathcal{V}) \), given by:
\begin{equation}
    (K_{V|S} f')(v) = \int p_{V|S}(v | s) f'(s) \, ds.
\end{equation}
Similarly, let \( T \) be a random variable with support \( \mathcal{T} \). A kernel operator \( K_{T;V | S} \) is defined as a mapping from a function \( f' \) in a function space \( \mathcal{F}(\mathcal{S}) \) onto a function \( K_{T;V | S} f' \) in \( \mathcal{F}(\mathcal{V}) \), given by:
\begin{equation}
    (K_{T;V | S} f')(v) = \int p_{T,V | S}(t, v | s) f'(s) \, ds.
\end{equation}
Moreover, the scaling operator \( \Lambda_{V | S} \) maps the function \( f'(s) \) to another function \( (\Lambda_{V | S} f')(s) \) defined by the pointwise multiplication as follows:
\begin{equation}
    (\Lambda_{V | S} f')(s) = p_{V | S}(v | s) f'(s).
\end{equation}
Starting from the kernel operator \( K_{X_C;X_A|X_B} \), defined as:
\begin{align}
    \big[K_{X_C;X_A|X_B} f'\big](x_A) &= \int p_{X_C X_A | X_B}(x_C, x_A | x_B) f'(x_B) \, dx_B,
\end{align}
we substitute the decomposition of \( p_{X_C X_A | X_B} \) as obtained previously:
\begin{align*}
    \big[K_{X_C;X_A|X_B} f'\big](x_A) &= \int \int p_{X_A | Z}(x_A | z) p_{X_C | Z}(x_C | z) \, p_{Z | X_B}(z | x_B) f'(x_B) \, dx_B \, dz.
\end{align*}
Rearranging the terms, we factorize the integral as:
\begin{align}
    \big[K_{X_C;X_A|X_B} f'\big](x_A) &= \int p_{X_A | Z}(x_A | z) \, p_{X_C | Z}(x_C | z) \, \big[K_{Z|X_B} f'\big](z) \, dz,
\end{align}
where the operator \( K_{Z|X_B} \) is defined as:
\begin{align}
    \big[K_{Z|X_B} f'\big](z) = \int p_{Z | X_B}(z | x_B) f'(x_B) \, dx_B.
\end{align}
Substituting $\Lambda_{X_C;Z}$ into the integral, we rewrite the equation as:
\begin{align}
    \big[K_{X_C;X_A|X_B} f'\big](x_A) &= \int p_{X_A | Z}(x_A | z) \, \big[\Lambda_{X_C;Z} K_{Z|X_B} f'\big](z) \, dz.
\end{align}
Finally, we apply the kernel operator \( K_{X_A|Z} \), defined as:
\begin{align}
    \big[K_{X_A|Z} f\big](x_A) = \int p_{X_A | Z}(x_A | z) f(z) \, dz,
\end{align}
to yield the operator equivalence:
\begin{align} \label{eq:operator_equivalence_final}
    \big[K_{X_C;X_A|X_B} f'\big](x_A) &= \big[K_{X_A|Z} \Lambda_{X_C;Z} K_{Z|X_B} f'\big](x_A).
\end{align}
This demonstrates the hierarchical decomposition of the operator \( K_{X_C;X_A|X_B} \) into a composition of the kernel operators \( K_{X_A|Z} \), \( K_{Z|X_B} \), and the scaling operator \( \Lambda_{X_C;Z} \), reflecting the conditional independence structure of the observed variables.

From Equation \eqref{eq:operator_equivalence_final}, we derive the operator equivalence:
\begin{equation}
    K_{X_C;X_A|X_B} = K_{X_A|Z} \Lambda_{X_C;Z} K_{Z|X_B}. \label{eq:operator_equivalence_rewrite_1}
\end{equation}
This equivalence holds over the space of functions \( \mathcal{G}(\mathcal{Z}) \), given the factorization properties of the conditional densities established earlier.

Now, integrating over \( X_C \), we use the fact that:
\begin{equation}
\int K_{X_C;X_A|X_B} f'(x_C) \, dx_C = K_{X_A|X_B} f',
\end{equation}
and for the scaling operator:
\begin{equation}
\int \Lambda_{X_C;Z} f''(z) \, dx_C = f''(z),
\end{equation}
which together imply:
\begin{equation}
    K_{X_A|X_B} = K_{X_A|Z} K_{Z|X_B}. \label{eq:operator_equivalence_rewrite_2}
\end{equation}

For any two functions \( f_1, f_2 \in L^2(\mathbb{Z}) \) satisfy \( K_{X_A|Z} f_1(x_A) = K_{X_A|Z} f_2(x_A) \) for all \( x_A \in X_A \). Then:

\begin{equation}
K_{X_A|Z} (f_1 - f_2)(x_A) = \int p_{X_A|Z}(x_A|z)(f_1(z) - f_2(z))dz = 0 , \forall x_A.
\end{equation}

Assumption \ref{assumption 3.2} implies that if \( \alpha_1(z) \) (here \( \alpha_1 = f_1 - f_2 \)) satisfies:

\begin{equation}
\int p_{X_A|Z}(x_A|z) \alpha_1(z) dz = 0 \text{ for all } x_A \in X_A,
\end{equation}

then \( \alpha_1(z) = 0 \) almost everywhere in \( Z \). Therefore, \( f_1(z) - f_2(z) = 0 \) almost everywhere in \( Z \), which means \( f_1 = f_2 \) in \( L^2(Z) \).

Thus, \(  K_{X_A|Z} \) is injective. Using this property, we deduce that \( K_{Z|X_B} \) can be expressed as:
\begin{equation}
    K_{Z|X_B} = K_{X_A|Z}^{-1} K_{X_A|X_B}. \label{eq:operator_equivalence_rewrite_3}
\end{equation}

The well-definedness of \( K_{X_A|Z}^{-1} \) over a sufficiently large domain ensures that this operator equivalence holds consistently. Substituting the expression for \( K_{Z|X_B} \) from Equation \eqref{eq:operator_equivalence_rewrite_3} into Equation \eqref{eq:operator_equivalence_rewrite_1}, we arrive at:
\begin{equation}
    K_{X_C;X_A|X_B} = K_{X_A|Z} \Lambda_{X_C;Z} K_{X_A|Z}^{-1} K_{X_A|X_B}. \label{eq:operator_equivalence_rewrite_4}
\end{equation}

The injectivity of \( K_{X_A|X_B} \) can be established by examining its adjoint operator \( K_{X_A|X_B}^\dagger \). Under Assumption \ref{assumption 3.2}, injectivity of \( K_{X_B|X_A} \) implies injectivity of \( K_{X_A|X_B}^\dagger \), which is the adjoint operator of $K_{X_A|X_B}$, since for any \( g(\cdot)/f_{X_A}(\cdot) \in \mathcal{F}(\mathcal{X}_A) \), the condition \( g \in \mathcal{F}(\mathcal{X}_A) \) holds.

We then view $K_{X_A|X_B}$ as a mapping of the closure of the range \( \mathcal{R}(K_{X_A|X_B}^\dagger) \) into \( \mathcal{F}(\mathcal{X}_A) \). By the closed range theorem, the closure \( \overline{\mathcal{R}(K_{X_A|X_B}^\dagger)} \) is the orthogonal complement of the null space of \( K_{X_A|X_B} \), denoted \( \mathcal{N}(K_{X_A|X_B}) \), and \(  \overline{\mathcal{R}(K_{X_A|X_B})} \) is the orthogonal complement of \( \mathcal{N}(K_{X_A|X_B}^\dagger) \). Therefore, $K^{-1}_{X_A|X_B}$ exists.

Because \( K_{X_A|X_B}^\dagger \) is injective, we have \( \mathcal{N}(K_{X_A|X_B}^\dagger) = \{0\} \). Consequently, \( \overline{\mathcal{R}(K_{X_A|X_B})} = \mathcal{F}(\mathcal{X}_A) \), and the inverse \( K_{X_A|X_B}^{-1} \) is well-defined and densely defined over \( \mathcal{F}(\mathcal{X}_A) \).

This result allows us to write:
\[
    K_{X_C;X_A|X_B} K_{X_A|X_B}^{-1} = K_{X_A|Z} \Lambda_{X_C;Z} K_{X_A|Z}^{-1}.
\]

The operator \( K_{X_C;X_A|X_B} K_{X_A|X_B}^{-1} \) admits a spectral decomposition, where the eigenvalues are given by the diagonal elements of \( \Lambda_{X_C;Z} \), i.e., \( \{p_{X_C|Z}(x_C | z)\} \), and the eigenfunctions are given by the kernel of \( K_{X_A|Z} \), i.e., \( \{p_{X_A|Z}(x_A | z)\} \).

Finally, the identification of \( p_{X_C|Z}(x_C|z) \) and \( p_{X_A|Z}(\cdot|z) \) is guaranteed by the uniqueness of the spectral decomposition, which is ensured by the injectivity of \( K_{X_A|X_B} \) and \( K_{X_A|Z} \).

Since \(K_{X_A|X_B}\) is injective, its inverse \(K_{X_A|X_B}^{-1}\) exists.
Substituting into Equation~\eqref{eq:operator_equivalence_rewrite_4}, we obtain
\begin{equation}\label{eq:decomposition_svd}
  K_{X_C;X_A|X_B}\,K_{X_A|X_B}^{-1}
  \;=\;
  K_{X_A|Z}\,\Lambda_{X_C;Z}\,K_{X_A|Z}^{-1}.
\end{equation}
Define
\[
  T
  \;\coloneqq\;
  K_{X_C;X_A|X_B}\,K_{X_A|X_B}^{-1}.
\]
By Assumption~\ref{assumption 3.1}, the conditional densities are bounded, so \(T\) is a bounded operator.
Moreover, the structure of \(T\) implies it admits a spectral decomposition in which the eigenvalues of \(T\) are precisely the entries of \(\Lambda_{X_C;Z}\), i.e., \(\{p_{X_C|Z}(x_C\mid z)\}\), and the corresponding eigenfunctions are encoded in the columns of \(K_{X_A|Z}\), i.e., \(\{p_{X_A|Z}(x_A\mid z)\}\). 

Hence, by the uniqueness of the spectral measure associated with such an operator (see e.g., \cite[Ch.~VII]{conway2019course}),
the decomposition into eigenvalues and eigenfunctions can only be realized in one way, up to standard indeterminacies. The first indeterminacy is the scaling. Specifically, we could replace $K_{X_A|Z}$ by $\alpha\,K_{X_A|Z}$ and $K_{X_A|Z}^{-1}$ by $(1/\alpha)\,K_{X_A|Z}^{-1}$ for any nonzero $\alpha$, leaving $T$ unchanged. But because $p_{X_C|Z}$ is a conditional density satisfying
\begin{equation}
\int p_{X_C|Z}(x_c|z)\,d x_c = 1,
\end{equation}
this normalizing condition forces $\alpha=1$. Hence, no further scaling is possible.

Another indeterminacy is due to the degeneracy of eigenvalues. The diagonal operator $\Lambda_{X_C;Z}$ has eigenvalues governed by $p_{X_C|Z}$. Clearly, without additional constraints, we could have distinct $Z$ values leading to the same eigenvalue. However, Assumption \ref{assumption 3.3} avoids this by ensuring the set $\left\{x : p(x_C \mid z) \neq p(x_C \mid z') \right\}$ has positive probability for all $z, z' \in \mathcal{Z}$ with $z \neq z'$.

Even if the eigenvalues are distinct, one can permute the labeling via a bijection $h : \mathcal{Z} \to \mathcal{Z}$. Instead of indexing by $Z$, we could reindex by $\tilde{Z} = h(Z)$, leaving $\Lambda$ unchanged but altering the labeling of $\Lambda_{X_C;Z}$. Consequently, the latent variable $Z$ is identified up to an invertible mapping $h$.  
\end{proof}

\subsection{Proof of Lemma \ref{lem:reduc1d}} \label{sec:proof:lem_reduc1d}

\ReduceoneD*

\begin{proof}
This is a special case of the Borsuk-Ulam theorem \citep{lyusternik1930topological, borsuk1933drei}. We prove this by a contradiction. Suppose $g$ is continuous.  $v_i \in \mathbb{R}^n$ for $i=1,2,3$ be distinct and their convex combinations be in the domain of $g$. Furthermore, we require that $$v_3\neq (1-\lambda) v_1 + \lambda v_2$$ for any $\lambda \in(0,1)$. 

Given that $g$ is one-to-one, we have
\begin{equation}
g(v_1) - g(v_2) \neq 0
\end{equation}
Consider a function $g:[0,1] \rightarrow \mathbb{R}$ as follows:
\begin{eqnarray*}
t(\lambda) &=& g(v_a(\lambda)) - g(v_b(\lambda)) \\
v_a(\lambda) &=& (1-\lambda) v_1 + \lambda v_2 \\
v_b(\lambda) &=& [1-\lambda(1-\lambda)][\lambda v_1 + (1-\lambda) v_2] + \lambda(1-\lambda)v_3
\end{eqnarray*}
Because $g$ is continuous, $t$ is a continuous function with 
\begin{eqnarray*}
t(0) &=& g(v_1) - g(v_2) \neq 0\\
t(1) &=& g(v_2) - g(v_1) \neq 0\\
t(1) &=& - t(0)
\end{eqnarray*}
Therefore, there must exist a $\lambda_0 \in(0,1)$ such that 
$$t(\lambda_0) =0$$
which means 
$$g(v_a(\lambda_0)) = g(v_b(\lambda_0)).$$
Because $v_a(\lambda_0) \neq v_b(\lambda_0)$, this is contradictory to the assumption that $g$ is one-to-one. Therefore, $g$ cannot be continuous.    
\end{proof}

\subsection{Proof of Lemma \ref{lem:reduckd}} \label{sec:proof:lem_reduckd}

\ReduceKD*

\begin{proof}
The Borsuk–Ulam theorem \citep{lyusternik1930topological, borsuk1933drei} states that if $g: S^k \rightarrow \mathbb{R}^k$ is continuous then there exists an $x \in S^k$ such that $g(-x)=g(x)$, where 
$$S^k= \{x \in \mathbb{R}^{k+1}: ||x||=1 \}.$$

Given that $k \leq n-1$, we show by contradiction that $g$ cannot be continuous and one-to-one, in particular, over $S^k$, a subset of the domain $\mathbb{R}^n$. Suppose $g$ is continuous. The Borsuk–Ulam theorem implies that there exists an $x \in S^k$ such that $g(-x)=g(x)$. That is contradictory to the assumption that $g$ is one-to-one. Therefore, $g$ cannot be continuous.
\end{proof}

\subsection{Proof of Theorem \ref{thm:sparsity}} \label{sec:proof:thm_sparsity}

\sparsity*

\begin{proof}
By Theorem \ref{thm:nf_suff}, there exists an invertible function $t$ such that $\hat{Z} = t(Z)$. Applying the chain rule to this transformation, we obtain the relationship between the Jacobians of $\hat{f}$ and $f$:
\begin{equation}
    J_{\hat{f}} = J_f J_t.
\end{equation}
Our goal is to show that $t$ is a composition of a permutation and component-wise transformations, which is equivalent to proving that $J_t$ is a generalized permutation matrix.

Let us denote $J_t$ by $\mathbf{T}$. According to our assumption, for each index $i$, the set of vectors $\{ J_f(z^{(\ell)})_{i,:} \}_{\ell=1}^{|\mathcal{F}_{i,:}|}$ spans the space $\mathbb{R}_{\mathcal{F}_{i,:}}^{n}$. This means any vector in $\mathbb{R}_{\mathcal{F}_{i,:}}^{n}$ can be expressed as a linear combination of these vectors. Specifically, for any standard basis vector $e_{j_0}$ (where $j_0 \in \mathcal{F}_{i,:}$), there exist coefficients $\alpha_{\ell}$ such that:
\begin{equation}
    e_{j_0} = \sum_{\ell \in \mathcal{F}_{i,:}} \alpha_{\ell} J_f(z^{(\ell)})_{i,:}.
\end{equation}
Multiplying both sides by $\mathrm{T}$, we get:
\begin{equation}
    e_{j_0} \mathrm{T} = \sum_{\ell \in \mathcal{F}_{i,:}} \alpha_{\ell} J_f(z^{(\ell)})_{i,:} \mathrm{T}.
\end{equation}
Since $J_{\hat{f}} = J_f \mathbf{T}$ and based on our assumptions, each term $J_f(z^{(\ell)})_{i,:} \mathrm{T}$ lies in $\mathbb{R}_{\hat{\mathcal{F}}_{i,:}}^{n}$. Therefore, $e_{j_0} \mathrm{T}$ also belongs to $\mathbb{R}_{\hat{\mathcal{F}}_{i,:}}^{n}$, implying:
\begin{equation}
    \mathrm{T}_{j_0,:} \in \mathbb{R}_{\hat{\mathcal{F}}_{i,:}}^{n}.
\end{equation}
This holds for all $j \in \mathcal{F}_{i,:}$, so we have:
\begin{equation}
    \forall j \in \mathcal{F}_{i,:},\quad \mathrm{T}_{j,:} \in \mathbb{R}_{\hat{\mathcal{F}}_{i,:}}^{n}.
\end{equation}
This establishes a connection between the supports:
\begin{equation} \label{eq:uc_connection_t5}
    \forall (i, j) \in \mathcal{F},\quad \{i\} \times \operatorname{supp}(\mathbf{T}_{j,:}) \subset \hat{\mathcal{F}}.
\end{equation}
A similar approach has been utilized in prior works such as \citep{strang2016introduction, lachapelle2021disentanglement, zhengidentifiability}. Since $J_f(z^{(\ell)})$ and $J_{\hat{f}}(\hat{z}^{(\ell)})$ both have full column rank $n$, the matrix $\mathbf{T}(z^{(\ell)})$ must be invertible, meaning its determinant is non-zero. Using the Leibniz formula for the determinant:
\begin{equation}
    \det(\mathbf{T}(z^{(\ell)})) = \sum_{\sigma \in \mathcal{S}_n} \operatorname{sgn}(\sigma) \prod_{i=1}^{n} \mathbf{T}(z^{(\ell)})_{i, \sigma(i)} \neq 0,
\end{equation}
where $\mathcal{S}_n$ is the set of all permutations of $\{1, \ldots, n\}$. Therefore, there exists at least one permutation $\sigma$ such that:
\begin{equation}
    \forall i \in \{1, \ldots, n\},\quad \mathbf{T}(z^{(\ell)})_{i, \sigma(i)} \neq 0.
\end{equation}
This implies that $\sigma(j) \in \operatorname{supp}(\mathbf{T}_{j,:})$ for all $j$. Combining this with Eq. \eqref{eq:uc_connection_t5}, we obtain:
\begin{equation}
    \forall (i, j) \in \mathcal{F},\quad (i, \sigma(j)) \in \hat{\mathcal{F}}.
\end{equation}
Define the permuted set:
\begin{equation}
    \sigma(\mathcal{F}) = \{ (i, \sigma(j)) \mid (i, j) \in \mathcal{F} \}.
\end{equation}
Thus, we have:
\begin{equation} \label{eq:uc_inclusion_t5}
    \sigma(\mathcal{F}) \subset \hat{\mathcal{F}}.
\end{equation}
Due to sparsity regularization on the estimated Jacobian, we know:
\begin{equation}
    |\hat{\mathcal{F}}| \leq |\mathcal{F}| = |\sigma(\mathcal{F})|.
\end{equation}
Combining this with Eq. \eqref{eq:uc_inclusion_t5}, it follows that:
\begin{equation} \label{eq:uc_equality_t5}
    \sigma(\mathcal{F}) = \hat{\mathcal{F}}.
\end{equation}

Suppose, for the sake of contradiction, that $\mathbf{T}(z)$ is not a composition of a diagonal matrix and a permutation matrix, i.e., there exist $j_1 \neq j_2$ such that:
\begin{equation}
    \operatorname{supp}(\mathbf{T}_{j_1,:}) \cap \operatorname{supp}(\mathbf{T}_{j_2,:}) \neq \emptyset.
\end{equation}
Let $j_3$ be an element in this intersection, so $\sigma(j_3) \in \operatorname{supp}(\mathbf{T}_{j_1,:}) \cap \operatorname{supp}(\mathbf{T}_{j_2,:})$. Without loss of generality, assume $j_3 \neq j_1$. According to Assumption \ref{assum:sparsity}, there exists a set $\mathcal{C}_{j_1}$ containing $j_1$ such that:
\begin{equation}
    \bigcap_{i \in \mathcal{C}_{j_1}} \mathcal{F}_{i,:} = \{ j_1 \}.
\end{equation}
Since $j_3 \neq j_1$, it must be that:
\begin{equation}
    j_3 \notin \bigcap_{i \in \mathcal{C}_{j_1}} \mathcal{F}_{i,:},
\end{equation}
implying there exists some $i_3 \in \mathcal{C}_{j_1}$ such that:
\begin{equation} \label{eq:uc_contradiction_t5}
    j_3 \notin \mathcal{F}_{i_3,:}.
\end{equation}
However, since $j_1 \in \mathcal{F}_{i_3,:}$, we have $(i_3, j_1) \in \mathcal{F}$. Using Eq. \eqref{eq:uc_connection_t5}, we find:
\begin{equation}
    (i_3, \sigma(j_3)) \in \hat{\mathcal{F}}.
\end{equation}
But from Eq. \eqref{eq:uc_equality_t5}, this means $(i_3, j_3) \in \mathcal{F}$, which contradicts Eq. \eqref{eq:uc_contradiction_t5}. This contradiction implies our assumption is false, and therefore $\mathbf{T}(z)$ must be a composition of a permutation matrix and a diagonal matrix.

Together with the equation $J_{\hat{f}} = J_f \mathbf{T}$, we achieve the desired result that $t$ is composed of a permutation and component-wise invertible functions.
\end{proof}

\subsection{Proof of Theorem \ref{thm:change}} \label{sec:proof:thm_change}

\change*

\begin{proof}
By Theorem \ref{thm:nf_suff}, there exists an invertible function $t$ such that $\hat{Z} = t(Z)$. Applying the change of variables formula for conditional densities, we have:
\begin{equation} \label{eq:change_of_variables}
    p_{\hat{Z} | U}(\hat{z} | U) = p_{Z | U}(z | U) \left| \det\left( J_{t^{-1}}(\hat{z}) \right) \right|.
\end{equation}

Taking the logarithm of both sides:
\begin{equation} \label{eq:log_densities}
    \log p_{\hat{Z} | U}(\hat{z} | U) = \log p_{Z | U}(z | U) + \log \left| \det\left( J_{t^{-1}}(\hat{z}) \right) \right|.
\end{equation}

Assuming that the conditional density $p_{Z | U}(z | U)$ factorizes over components, i.e.,
\begin{equation} \label{eq:factorization}
    p_{Z | U}(z | U) = \prod_{i=1}^{n} p_{Z_i | U}(z_i | U),
\end{equation}
and similarly for $p_{\hat{Z} | U}(\hat{z} | U)$. Substituting Eq.~\eqref{eq:factorization} into Eq.~\eqref{eq:log_densities}, we obtain:
\begin{equation} \label{eq:log_densities_expanded}
    \sum_{i=1}^{n} \log p_{\hat{Z}_i | U}(\hat{z}_i | U) = \sum_{i=1}^{n} \log p_{Z_i | U}(z_i | U) + \log \left| \det\left( J_{t^{-1}}(\hat{z}) \right) \right|.
\end{equation}

Next, following a common technique in the literature \citep{hyvarinen2024identifiability}, we take the second derivatives of both sides with respect to $\hat{Z}_k$ and $\hat{Z}_v$, where $k \neq v$. Note that for $i \neq k$, we have $\partial \log p_{\hat{Z}_i | U}(\hat{z}_i | U)/\partial \hat{Z}_k = 0$. Therefore, the left-hand side simplifies to:
\begin{equation} \label{eq:left_hand_side}
    \frac{\partial^2}{\partial \hat{Z}_k \partial \hat{Z}_v} \sum_{i=1}^{n} \log p_{\hat{Z}_i | U}(\hat{z}_i | U) = 0.
\end{equation}

For the right-hand side, we define:
\begin{align}
    h'_{i, (k)} &:= \frac{\partial Z_{i}}{\partial \hat{Z}_{k}}, \\
    h''_{i, (k, v)} &:= \frac{\partial^2 Z_{i}}{\partial \hat{Z}_{k} \partial \hat{Z}_{v}}, \\
    \eta'_i ( z_{i}, U ) &:= \frac{
    \partial \log p_{Z_i | U}(z_i | U) 
    }{
        \partial Z_{i}
    }, \\
    \eta''_i ( z_{i}, U ) &:= \frac{
    \partial^{2} \log p_{Z_i | U}(z_i | U)
    }{
        (\partial Z_{i})^{2}
    }.
\end{align}

Then, the second derivative of the right-hand side is:
\begin{equation}
    \sum_{i=1}^{n} \left( 
        \eta''_{i} ( z_{i}, U ) \cdot h'_{i, (k)} h'_{i, (v)}  
        + \eta'_{i} ( z_{i}, U ) \cdot  h''_{i, (k,v)} 
    \right) + \frac{\partial^2}{\partial \hat{z}_{k} \partial \hat{z}_{v}} \log \left| \det\left( J_{t^{-1}}(\hat{z}) \right) \right|.
\end{equation}

Setting the left-hand side and right-hand side equal and simplifying, we obtain:
\begin{equation} \label{eq:second_derivative_equation}
    \sum_{i=1}^{n} \left( 
        \eta''_{i} ( z_{i}, U ) \cdot h'_{i, (k)} h'_{i, (v)}  
        + \eta'_{i} ( z_{i}, U ) \cdot  h''_{i, (k,v)} 
    \right) + \frac{\partial^2}{\partial \hat{z}_{k} \partial \hat{z}_{v}} \log \left| \det\left( J_{t^{-1}}(\hat{z}) \right) \right| = 0.
\end{equation}

Consider $2n + 1$ different values of $U$, denoted by $U^{(i)}$ for $i \in \{0,1,\ldots,2n\}$. Evaluating Eq.~\eqref{eq:second_derivative_equation} at these values, we obtain $2n + 1$ equations. Subtracting the equation corresponding to $U^{(0)}$ from each of the other equations, we get $2n$ equations:
\begin{equation} \label{eq:linear_system}
    \sum_{i=1}^{n} \left( 
        \left[ \eta''_{i} ( z_{i}, U^{(j)} ) - \eta''_{i} ( z_{i}, U^{(0)} ) \right] h'_{i, (k)} h'_{i, (v)}  
        + \left[ \eta'_{i} ( z_{i}, U^{(j)} ) - \eta'_{i} ( z_{i}, U^{(0)} ) \right] h''_{i, (k,v)} 
    \right) = 0,
\end{equation}
for $j \in \{0,1,\ldots,2n\}$.

Under Assumption \ref{assum:change}, the $2n$ vectors formed by the differences $\mathbf{w}(Z, U^{(j)}) - \mathbf{w}(Z, U^{(0)})$ are linearly independent. This implies that the only solution to the linear system in Eq.~\eqref{eq:linear_system} is:
\begin{equation}
    h'_{i, (k)} h'_{i, (v)} = 0 \quad \text{and} \quad h''_{i, (k,v)} = 0, \quad \text{for all } i \text{ and } k \neq v.
\end{equation}

This means that for each $i$, there is at most one index $r_i$ such that $h'_{i, (r_i)} \neq 0$, and all mixed second derivatives $h''_{i, (k,v)}$ with $k \neq v$ are zero. Since $t^{-1}$ is invertible, each row of the Jacobian $J_{t^{-1}}(\hat{Z})$ must have at least one non-zero entry. Therefore, there exists a permutation $\pi$ such that each $Z_i$ depends only on $\hat{Z}_{\pi(i)}$, i.e.,
\begin{equation}
    Z_i = h_i^{-1}(\hat{Z}_{\pi(i)}).
\end{equation}
Equivalently, we have the following equation:
\begin{equation}
    \hat{Z}_{\pi(i)} = h_i(Z_i),
\end{equation}
where $h_i$ is a univariate invertible function.

Thus, $\hat{z}$ is related to $z$ through a component-wise invertible transformation composed with a permutation. This completes the proof.
\end{proof}

\section{Supplementary Experiments}
\label{appx:experiments}

\subsection{Supplementary details of the settings}

\textbf{Generating process for discrete data.} \ 
For the baseline case, we use 
\begin{equation*}
\begin{aligned}
    k = 4&, \quad  \epsilon_1 \sim \mathcal{N}(0, 1), \\
    f_1(z) = z&, \quad  \epsilon_2 \sim Beta(2, 2) - \frac{1}{2}, \\
    f_2(z) = \frac{1}{1+e^z}&, \quad  \epsilon_3 \sim Laplace(0, 1), \\
    f_3(z) = z^2&, \quad  \epsilon_4 \sim Bernoulli\left(\frac{1}{2}\right),\\
    X_4 = \Phi(Z/3) \cdot &(-1)^{I(\epsilon_4 > 0.5)},  Z \sim Binomial(10, 0.5).
\end{aligned}
\end{equation*}
For the linear error case, we use:  
\begin{equation*}
    \epsilon_1 =  \mathcal{N}(0, \frac{1}{4}Z^2), \epsilon_3 =  Laplace(0, \frac{1}{2}|Z|).
\end{equation*}
For the double error case, we use:  
\begin{equation*}
    \epsilon_1 =  \mathcal{N}(0, 4), \epsilon_2  =  Beta(2, 4), \epsilon_3  =  Laplace(0, 2).
\end{equation*}

\textbf{Additional details for basis validation.} \
We use a $6$-layer with $10$ hidden nodes fully connected neural network. The window size $w$ and normalization term $\lambda$ are tuned as hyper-parameters.  In the loss function defined in Eq, \eqref{equ loss}, it requires more than one data point to estimate the kernel density function. As a result, unlike other use cases that one training point is enough to calculate its corresponding loss, we need to sample $M$ ($>1$) points as one observation to calculate its loss. For example, to build the training sample we sample with replacement $M$ points from the entire training data points and repeat $N$ times, and we end up with $N$ observations in our training sample. We use kernel functions to approximate their density functions. The kernel function $K(\cdot)$ can simply be the standard normal density function. For the bandwidth, we adopt the so-called Silverman's rule, i.e., $h_j = w \sigma_j N^{-1/5}$ where $ \sigma_j$ is the standard error of $X_j$, and $w$ is the window size that is determined by hyper parameters tuning. Similarly, we may take $h^* = w \hat{\sigma} N^{-1/5}$, where $ \hat{\sigma}$ is the standard error of $\hat{Z}$. Theoretically, if a distribution is normal, the best choice for $w$ used in the kernel function is $1$, so to tune $w$ we choose the range from $0.5$ to $4$. To tune $\lambda$, we arbitrarily choose the range from $0$ to $1$.

\textbf{Additional details for generalized validation.} \
For all datasets used in generalized validation, the training set consists of $10,000$ samples, and the test set consists of $2,000$ samples. For datasets satisfying structural variability, each observed variable is generated through a transformation of its dependent latent variables based on the structural condition. For datasets satisfying distributional variability, latent variables are sampled from $2n+1$ Gaussian distributions. All experiments are repeated over 5 runs with different random seeds and are performed on $12$ CPUs.

For the results w.r.t. different standard deviations of the noise (Figure. \ref{fig:noise_level}), we fix the number of latent variables as $5$ and vary the standard deviations of the noise across $\{0.5,1,1.5,2,2.5\}$.

\subsection{Supplementary experimental results}

\begin{figure}[t]
    \centering
    \includegraphics[width=0.7\linewidth]{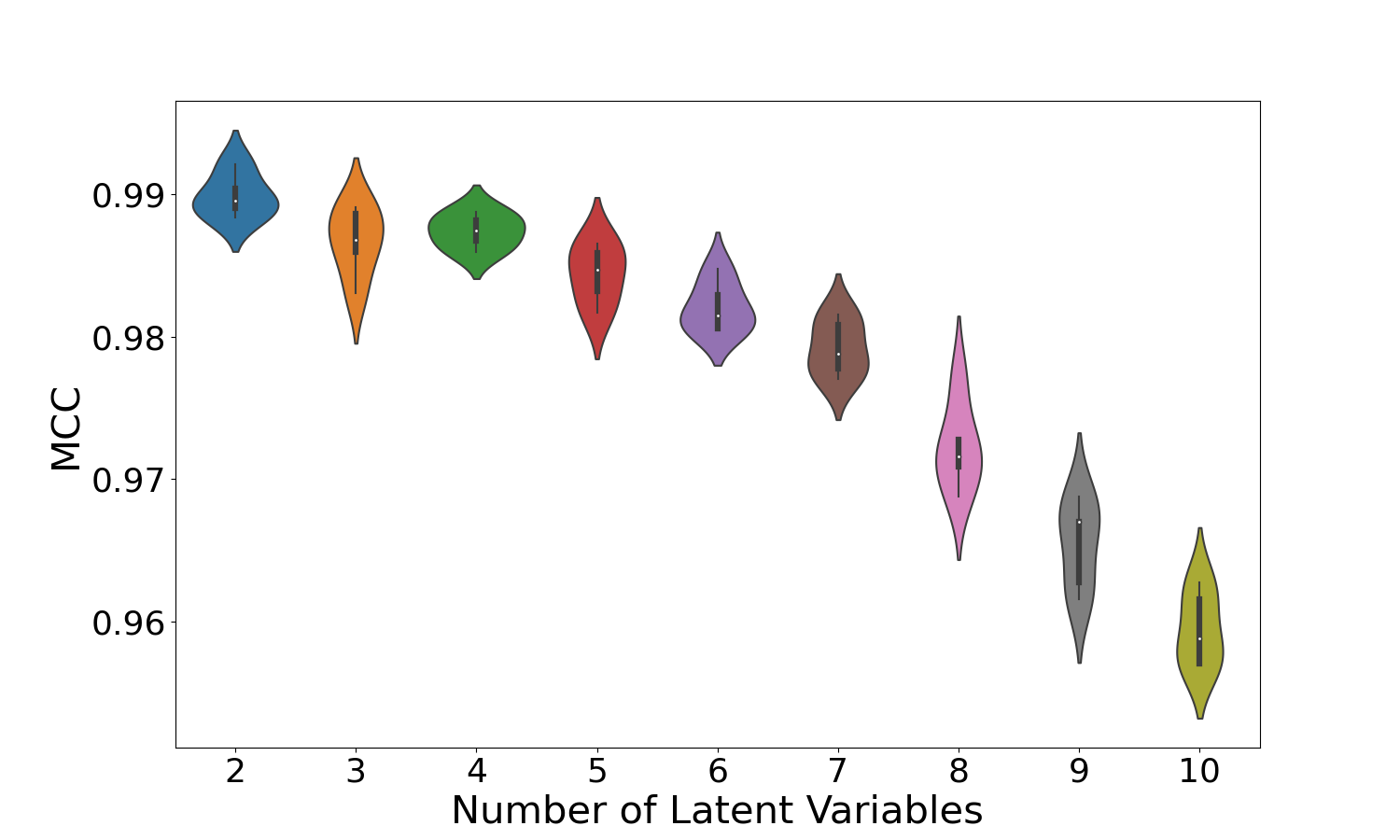}
    \caption{Results w.r.t. different numbers of latent variables for model satisfying distributional variability.}
    \vspace{-1em}
    \label{fig:latent_num_change}
\end{figure}

\begin{figure}[t]
    \centering
    \includegraphics[width=0.7\linewidth]{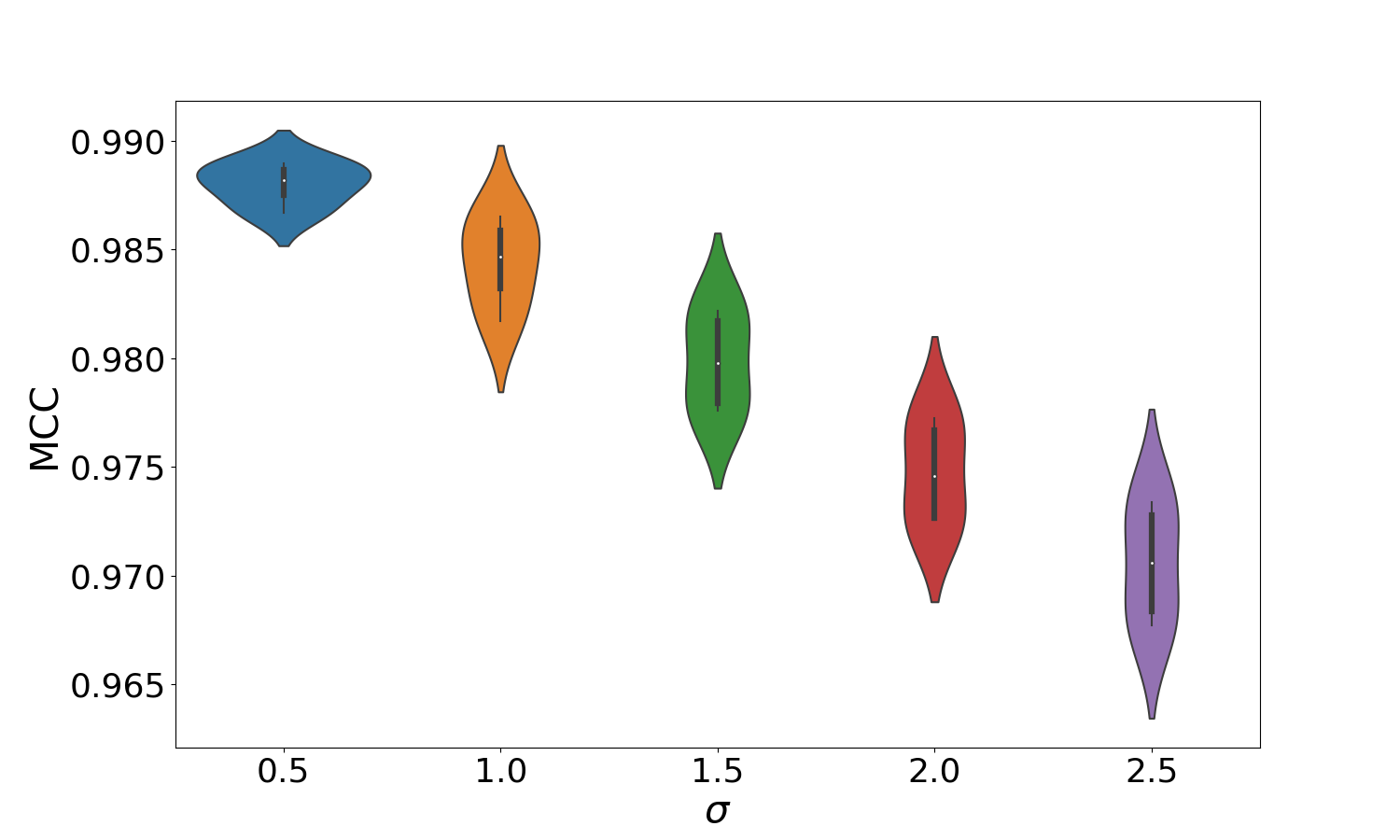}
    \caption{Results w.r.t. different standard deviations ($\sigma$) of the noise for model satisfying distributional variability. We set the number of latent variables $n$ as $5$.}
    \vspace{-1em}
    \label{fig:latent_noise_change}
\end{figure}

\textbf{Results with distributional variability.} \ 
In addition to the results with structural variability, we also validate our theory of the identifiability with distributional variability (Theorem \ref{thm:change}). For $n$ latent variables, we sample them from $2n+1$ Gaussian distributions, with means uniformly drawn from $[-5, 5]$ and variances from $[0.5, 2]$. The number of latent variables ranges from $2$ to $10$, and the observed variables are set as three times the corresponding latent variables.

The results are shown in Figure \ref{fig:latent_num_change}. We can observe that, across all settings, our model achieves high MCC consistently. This confirms the component-wise identifiability under the condition of distributional variability.

In addition, we also evaluate the model with different noise levels, i.e., different standard deviations of the noise. Specifically, we fix the number of latent variables as $5$ and vary the standard deviation across $\{0.5, 1, 1.5, 2, 2.5\}$. From the results (Figure \ref{fig:latent_noise_change}), we observe that the quality of identification stays robust across different noise levels. This further supports the proposed identifiability theory in complicated noisy settings.

\end{document}



% Supplementary material: To improve readability, you must use a single-column format for the supplementary material.
\onecolumn
\aistatstitle{Supplementary}

\section{FORMATTING INSTRUCTIONS}

To prepare a supplementary pdf file, we ask the authors to use \texttt{aistats2025.sty} as a style file and to follow the same formatting instructions as in the main paper.
The only difference is that the supplementary material must be in a \emph{single-column} format.
You can use \texttt{supplement.tex} in our starter pack as a starting point, or append the supplementary content to the main paper and split the final PDF into two separate files.

Note that reviewers are under no obligation to examine your supplementary material.

\section{MISSING PROOFS}

The supplementary materials may contain detailed proofs of the results that are missing in the main paper.

\subsection{Proof of Lemma 3}

\textit{In this section, we present the detailed proof of Lemma 3 and then [ ... ]}

\section{ADDITIONAL EXPERIMENTS}

If you have additional experimental results, you may include them in the supplementary materials.

\subsection{The Effect of Regularization Parameter}

\textit{Our algorithm depends on the regularization parameter $\lambda$. Figure 1 below illustrates the effect of this parameter on the performance of our algorithm. As we can see, [ ... ]}

\vfill